\documentclass[letterpaper,journal]{IEEEtran}
\usepackage{amsmath,amsfonts}
\usepackage{algorithmic}
\usepackage{algorithm}
\usepackage{array}
\usepackage[caption=false,font=normalsize,labelfont=sf,textfont=sf]{subfig}
\usepackage{textcomp}
\usepackage{stfloats}
\usepackage{url}
\usepackage{verbatim}
\usepackage{graphicx}
\usepackage{cite}
\usepackage{orcidlink}

\usepackage{pifont}
\usepackage{colortbl}
\usepackage{graphicx}
\usepackage{amsmath}
\usepackage{amssymb}
\usepackage{booktabs}
\usepackage{enumitem}
\usepackage{booktabs, multirow} 

\definecolor{ceiling}{RGB}{214,  38, 40}   %
\definecolor{floor}{RGB}{43, 160, 4}     %
\definecolor{wall}{RGB}{158, 216, 229}  %
\definecolor{window}{RGB}{114, 158, 206}  %
\definecolor{chair}{RGB}{204, 204, 91}   %
\definecolor{bed}{RGB}{255, 186, 119}  %
\definecolor{sofa}{RGB}{147, 102, 188}  %
\definecolor{table}{RGB}{30, 119, 181}   %
\definecolor{tvs}{RGB}{160, 188, 33}   %
\definecolor{furniture}{RGB}{255, 127, 12}  %
\definecolor{objects}{RGB}{196, 175, 214} %
\definecolor{ForestGreen}{RGB}{34, 139, 34}

\usepackage{xcolor}  
\usepackage{utfsym}  

\makeatletter
\@ifpackageloaded{xcolor}{}{\usepackage[table]{xcolor}}
\makeatother

\usepackage{amsmath}
\usepackage{amssymb}
\IfFileExists{mathrsfs.sty}{\usepackage{mathrsfs}}{}
\IfFileExists{bbm.sty}{\usepackage{bbm}}{}

\usepackage{graphicx}
\IfFileExists{adjustbox.sty}{\usepackage{adjustbox}}{}
\usepackage{tabularx}
\IfFileExists{tabularray.sty}{\usepackage{tabularray}}{}
\usepackage{array}
\usepackage{multirow}
\IfFileExists{makecell.sty}{\usepackage{makecell}}{}
\IfFileExists{wrapfig.sty}{\usepackage{wrapfig}}{}
\usepackage{booktabs}
\usepackage{colortbl}

\usepackage{algorithm}
\makeatletter
\@ifpackageloaded{algorithmic}{}{\IfFileExists{algpseudocode.sty}{\usepackage{algpseudocode}}{}}
\makeatother

\IfFileExists{xspace.sty}{\usepackage{xspace}}{}
\IfFileExists{utfsym.sty}{\usepackage{utfsym}}{}
\IfFileExists{marvosym.sty}{\usepackage{marvosym}}{}
\usepackage{enumitem}
\usepackage{pifont}
\IfFileExists{pgffor.sty}{\usepackage{pgffor}}{}
\IfFileExists{svg.sty}{\usepackage{svg}}{}

\providecommand{\bh}[1]{\textcolor{black}{#1}}

\providecommand{\Letter}{\ding{41}}

\definecolor{occ_ceiling}{RGB}{214,38,40}
\definecolor{occ_floor}{RGB}{43,160,4}
\definecolor{occ_wall}{RGB}{158,216,229}
\definecolor{occ_window}{RGB}{114,158,206}
\definecolor{occ_chair}{RGB}{204,204,91}
\definecolor{occ_bed}{RGB}{255,186,119}
\definecolor{occ_sofa}{RGB}{147,102,188}
\definecolor{occ_table}{RGB}{30,119,181}
\definecolor{occ_tvs}{RGB}{160,188,33}
\definecolor{occ_furniture}{RGB}{255,127,12}
\definecolor{occ_objects}{RGB}{196,175,214}

\definecolor{colorfirst}{rgb}{.866,.945,0.831}
\definecolor{colorsecond}{rgb}{1,0.98,0.83}
\definecolor{colorthird}{rgb}{0.76,0.87,0.92}
\definecolor{colorcite}{rgb}{0.212,0.490,0.741}
\definecolor{occ_lightgray}{gray}{0.92}

\makeatletter
\providecommand{\onedot}{\futurelet\@let@token\@onedot}
\providecommand{\@onedot}{\ifx\@let@token.\else.\null\fi\xspace}

\makeatother

\begin{document}

\title{
Bridging 3D Gaussians and Semantic Occupancy for Comprehensive Open-Vocabulary Scene Understanding from Unposed Images
}

\author{
~Hu Zhu$^*$,~Bohan Li$^*$,~Xianda Guo$^*$,~Yanlun Peng,~Zheng Zhu,\\~Xin Jin,~Wenjun Zeng\textsuperscript{\Letter},~\IEEEmembership{Fellow, IEEE} and Chang Wen Chen,~\IEEEmembership{Life Fellow, IEEE}
\thanks{Hu Zhu is with the Department of Computing, The Hong Kong Polytechnic University, Hong Kong, SAR, China, and Eastern Institute of Technology, Ningbo, China. Bohan Li is with Shanghai Jiao Tong University, Shanghai, China, and Eastern Institute of Technology, Ningbo, China. Xianda Guo is with Wuhan University, Wuhan, China. 
Yanlun Peng is with Great Wall Motor, China. Zheng Zhu is with Tsinghua University, Beijing, China.
Hu Zhu, Bohan Li and Xianda Guo share equal contribution.
}
\thanks{Wenjun Zeng~(corresponding author) is a chair professor, and Xin Jin is an assistant professor at the Ningbo Institute of Digital Twin, Eastern Institute of Technology, Ningbo, China.}	
\thanks{Chang Wen Chen is with the Department of Computing, The Hong Kong Polytechnic University, Hong Kong, SAR, China~(e-mail: changwen.chen@polyu.edu.hk).}
}


\maketitle

\begin{abstract}
Comprehensive 3D scene understanding from sparse, unposed images requires a model to recover renderable geometry, open-vocabulary semantics, and free/occupied 3D space without relying on external camera calibration. Recent feed-forward Gaussian methods improve pose-free reconstruction and semantic rendering, but their Gaussian primitives are mainly optimized through image-space objectives and remain weakly constrained in unobserved regions. We propose \textit{COVScene}, a pose-free semantic Gaussian framework that couples renderable Gaussian primitives with a dense semantic occupancy field through differentiable volumetric lifting. Instead of converting Gaussians to voxels only at evaluation time, COVScene lifts the predicted semantic Gaussians inside the training computation graph, so volumetric regularization provides gradients to Gaussian opacity, geometry, and semantic features. The framework combines a semantic-aware Geometry Transformer, multi-task Gaussian decoding, geometric foundation distillation, and occupancy entropy regularization to support novel view synthesis, open-vocabulary semantic querying, and semantic occupancy prediction within a single representation. Experiments on ScanNet and ScanNet++ show that COVScene maintains competitive rendering quality, improves open-vocabulary segmentation, and achieves stronger semantic occupancy prediction than the self-supervised baseline without direct voxel-level supervision.\looseness=-1

\end{abstract}

\begin{IEEEkeywords}
Open-Vocabulary 3D Scene Understanding, Unified Gaussian Splatting, Semantic Occupancy, Differentiable Volumetric Lifting
\end{IEEEkeywords}

\section{Introduction}

Comprehensive 3D scene understanding is an important problem in multimedia perception, requiring models to recover visual semantics, geometry, and physical space occupancy. Recent studies have advanced this goal through camera-based 3D semantic scene completion~\cite{xuepinet2026,li2023from}, geometry-aware image-based 3D perception~\cite{zhang20253dgeo,jia2022conv}, multimodal 3D scene reasoning~\cite{xiong20253urllm}, and open-vocabulary visual understanding~\cite{wang2026adapt,zhang2026detagent,zhang2025unleash,fang2026casovd}. This paper focuses on a more unconstrained setting: robust 3D scene understanding from sparse, unposed images. For robotics, embodied agents, and augmented reality, a reconstructed scene should support novel-view synthesis, open-vocabulary semantic querying, and explicit reasoning about which regions of 3D space are free or occupied. Recent advances in Neural Radiance Fields (NeRF)~\cite{nerf} and 3D Gaussian Splatting (3DGS)~\cite{3dgs} have greatly improved renderable scene representations, yet many reconstruction pipelines still assume a known calibrated camera or camera poses estimated by Structure-from-Motion (SfM)~\cite{schonberger2016sfm}. Moreover, rendering-oriented representations alone do not directly provide a physically grounded volumetric description of the scene.\looseness=-1

Recent feed-forward reconstruction methods have made important progress toward scalable 3D reconstruction from sparse or uncalibrated images. Generalizable gaussian methods such as pixelSplat~\cite{charatan2024pixelsplat}, MVSplat~\cite{chen2024mvsplat}, and VolSplat~\cite{wang2025volsplat} learn image-to-gaussian priors for efficient sparse-view reconstruction, while unposed reconstruction models~\cite{ye2025noposplat,szymanowicz2024splatter,smart2024splatt3r,Wang2024dust3r,wang2025vggt,keetha2026mapanything} reduce or remove the need for external camera calibration. Building on this trend, pose-free Gaussian methods such as LSM~\cite{fan2024lsm}, AnySplat~\cite{jiang2025anysplat}, UniForward~\cite{tian2025uniforward}, and Uni3R~\cite{sun2025uni3r} predict renderable Gaussian representations from unposed multi-view inputs, and several of them further attach semantic features for open-vocabulary scene understanding. Concurrently, open-vocabulary 2D perception models such as LSeg~\cite{Li2022Lseg}, together with foundation models such as CLIP~\cite{clip} and SAM~\cite{kirillov2023sam}, have fostered a new wave of semantic 3D methods~\cite{guo2024semantic, zhou2024feature3dgs, fan2024lsm, shi2024language, wang2025language, sheng2025spatialsplat}. These approaches attempt to lift 2D semantic features into 3D space to achieve open-vocabulary understanding without extensive annotation. In particular, LSM~\cite{fan2024lsm} and Uni3R~\cite{sun2025uni3r} demonstrate that pose-free semantic gaussians can support novel-view synthesis, open-vocabulary segmentation, and depth prediction in a single feed-forward framework. These advances establish semantic Gaussian splatting as a strong representation for scalable 3D scene reconstruction.\looseness=-1

\begin{table*}[!t]
\centering
\caption{\textbf{Comparison of representative pose-free gaussian scene representations.} \ding{52} indicates supported; \ding{55} indicates not supported. `Post-voxel.' denotes Gaussian-to-occupancy conversion only at evaluation time, without volumetric feedback during training.}
\label{tab: capability_comparison}
\setlength{\tabcolsep}{6pt}
\renewcommand{\arraystretch}{1.2}
\resizebox{\textwidth}{!}{
\begin{tabular}{lcccccc}
\toprule
\textbf{Method} & \textbf{Pose-Free} & \textbf{Feed-Forward} & \textbf{Renderable GS} & \textbf{Open-Vocab. Sem.} & \textbf{Semantic Occ.} & \textbf{Train-Time Vol. Feedback} \\
\midrule
LSM~\cite{fan2024lsm} & \ding{52} & \ding{52} & \ding{52} & \ding{55} & \ding{55} & \ding{55} \\
AnySplat~\cite{jiang2025anysplat} & \ding{52} & \ding{52} & \ding{52} & \ding{55} & \ding{55} & \ding{55} \\
Uni3R~\cite{sun2025uni3r} & \ding{52} & \ding{52} & \ding{52} & \ding{52} & \ding{55} & \ding{55} \\
\rowcolor{blue!5} \textbf{COVScene} & \textbf{\ding{52}} & \textbf{\ding{52}} & \textbf{\ding{52}} & \textbf{\ding{52}} & \textbf{\ding{52}} & \textbf{\ding{52}} \\
\bottomrule
\end{tabular}
}
\end{table*}

Despite this progress, existing pose-free semantic Gaussian representations remain largely surface-centric. Their Gaussian opacity and semantic features are optimized via image-space rendering losses, but the learned primitives are not explicitly constrained to represent free and occupied 3D space. Although recent works improve Gaussian reconstruction efficiency or compactness~\cite{wang2024freesplat,sheng2025spatialsplat}, they do not enforce a physically plausible occupancy distribution during training. As summarized in Tab.~\ref{tab: capability_comparison}, a straightforward post-hoc voxelization of predicted Gaussians can expose a volumetric field during evaluation, but it cannot correct the underlying Gaussian representation without volumetric feedback during training. As a result, a model can produce plausible RGB or semantic renderings while still placing ambiguous density in unobserved regions, producing floaters, hollow structures, or physically implausible layouts.\looseness=-1

To address this limitation, we propose COVScene, an occupancy-grounded semantic Gaussian framework for pose-free 3D scene understanding. Rather than treating occupancy as an auxiliary branch or an offline conversion, COVScene differentiably lifts predicted semantic gaussians into a dense semantic occupancy field inside the training computation graph. The resulting volumetric field provides explicit free/occupied-space regularization, and its gradients directly update the same Gaussian opacity, geometry, and semantic features used for novel-view synthesis and open-vocabulary rendering. This closed-loop Gaussian-volume coupling makes the renderable representation more physically plausible while preserving the efficiency and flexibility of feed-forward semantic Gaussian reconstruction.\looseness=-1

We evaluate COVScene on novel-view synthesis, open-vocabulary segmentation, and semantic occupancy prediction. Beyond standard comparisons, we include controlled baselines based on semantic Gaussians, post-processing Gaussian voxelization, and ablated COVScene variants. These experiments show that simply converting semantic Gaussians to occupancy at evaluation is insufficient, whereas training-time volumetric grounding substantially improves semantic occupancy and reduces geometric artifacts, achieving competitive rendering and segmentation performance.

Our contributions are summarized as follows:

\begin{itemize}

\item We propose COVScene, an occupancy-grounded semantic Gaussian framework that couples renderable Gaussian primitives with a dense semantic occupancy field through differentiable volumetric lifting. The lifted field is used during training, rather than as detached post-processing, so volumetric losses directly regularize Gaussian opacity, geometry, and semantic features.

\item We introduce an occupancy regularization objective that encourages physically plausible free/occupied structure in weakly observed regions while preserving open-vocabulary rendering from the same Gaussian representation.

\item A multi-task adaptive decoder is designed to fuse depth features with probabilistic cost-volume cues. Crucially, an occupancy-entropy regularization strategy is introduced to enforce bimodal physical priors, thereby resolving geometric inconsistencies and artifacts inherent in previous unposed 3DGS.

\item  Extensive evaluations on ScanNet~\cite{scannet} and ScanNet++~\cite{scannetpp} demonstrate that COVScene achieves state-of-the-art (SOTA) performance across versatile tasks, including novel view synthesis, open-vocabulary semantic segmentation, and zero-shot 3D occupancy prediction.\looseness=-1
\end{itemize}

\begin{figure*}[!t]
  \centering
    \includegraphics[width=0.95\linewidth]{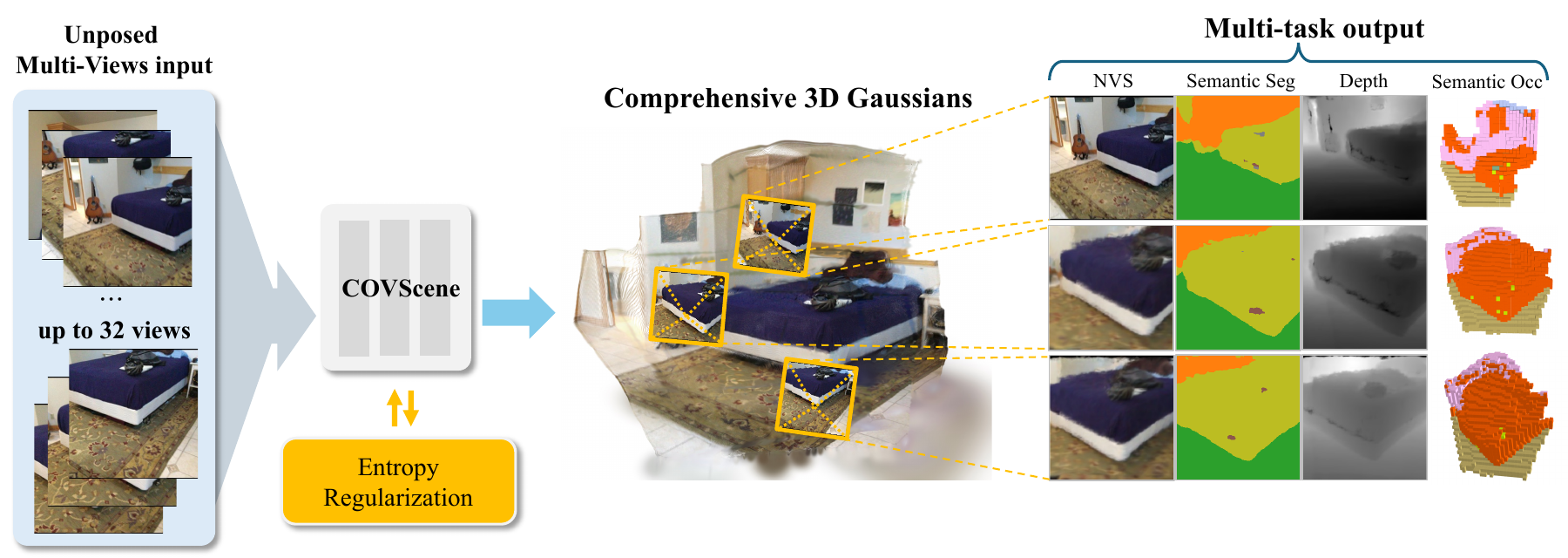}
    \vspace{-10pt}
  \caption{\textbf{Versatile capabilities of COVScene.} Given unposed multi-view images, a single forward pass directly constructs Comprehensive 3D Gaussians. This unified representation tightly couples appearance, geometry, and features, simultaneously enabling novel view synthesis, open-vocabulary semantic segmentation, depth estimation, and 3D semantic occupancy prediction.
  }
\label{fig: teaser}
\end{figure*}
 
\section{Related Work}
\subsection{Generalizable 3D Reconstruction Methods}
NeRF~\cite{nerf} and 3DGS~\cite{3dgs} have substantially advanced image-based 3D reconstruction and novel view synthesis~\cite{yu2020pixelnerf, zhuh2022fusing, ye2025noposplat, smart2024splatt3r}. However, many early radiance-field and Gaussian-based pipelines depend on per-scene optimization, which limits their efficiency and scalability. Subsequent acceleration techniques~\cite{mueller2022instantngp, kplanes_2023} reduce the optimization cost, but the reconstruction process still often requires iterative fitting for each scene. To improve generalization across scenes, feed-forward methods~\cite{charatan2024pixelsplat, chen2024mvsplat, yu2020pixelnerf} learn reconstruction priors from large-scale data and predict scene representations in a single network pass. In particular, PixelSplat~\cite{charatan2024pixelsplat} and MVSplat~\cite{chen2024mvsplat} use epipolar geometry~\cite{epipolartransformers} and cost volumes~\cite{yao2018mvsnet} to infer depth and Gaussian attributes from sparse image pairs. Later variants, including MVSplat360~\cite{chen2024mvsplat360}, DrivingForward~\cite{tian2025drivingforward}, and YoNoSplat~\cite{ye2026yonosplat}, extend feed-forward Gaussian reconstruction to broader camera distributions and more complex scenes. Another line of work reduces the dependence on known camera parameters. The DUST3R series~\cite{Wang2024dust3r, tang2025mvdust3r} estimates dense correspondences and camera geometry from image collections, and recent pose-free methods~\cite{ye2025noposplat, ye2025noposplat, szymanowicz2024splatter, smart2024splatt3r, zhang2025flare, hong2025pfplat} further couple camera estimation with feed-forward 3D reconstruction. These methods make reconstruction from unposed images increasingly practical, but most of them remain focused on geometry and rendering rather than dense semantic occupancy. COVScene follows the pose-free feed-forward setting, while introducing volumetric semantic occupancy as an explicit training-time constraint on the Gaussian representation.\looseness=-1

\vspace{-10pt}
\subsection{Open-Vocabulary 3D Scene Understanding}
Beyond geometric reconstruction, recent work has increasingly studied the semantic and structural understanding of 3D scenes~\cite{lerf2023,li2024one, garfield2024, li2023bridging, qin2024langsplat,li2025omninwm, li2025occscene, OpenSplat3D, li2026articulated,li2025uniscene,li2026hierarchical,li20254dlangsplat, konushin2025tun3d, peng2023openscene, zhang2025roboocc}. Early open-vocabulary 3D methods incorporate features from CLIP~\cite{clip} into NeRF fields, as demonstrated by LERF~\cite{lerf2023}. This idea has been extended to 3DGS to support efficient rendering and queryable semantic fields. For example, GARField~\cite{garfield2024} distills masks from SAM~\cite{kirillov2023sam} into Gaussian primitives, LangSplat~\cite{qin2024langsplat} constructs a scene-wise language field through feature auto-encoding, and Gaussian Grouping~\cite{gaussian_grouping} attaches identity codes for instance-level grouping. Further extensions, including OpenSplat3D~\cite{OpenSplat3D} and 4D LangSplat~\cite{li20254dlangsplat}, support promptable segmentation and temporally coherent language fields. OV-NeRF~\cite{liao2024ovnerf} improves cross-view semantic consistency through semantic field distillation, while MaskField~\cite{gao2024maskfield} decomposes SAM mask features in Gaussian Splatting for efficient 3D semantic segmentation. These methods demonstrate the value of lifting 2D semantic priors into 3D, but most of them assume calibrated inputs, rely on per-scene optimization, or remain centered on surface-level semantic fields.\looseness=-1

Recent feed-forward and pose-free methods move closer to scalable semantic 3D scene understanding. LSM~\cite{fan2024lsm} predicts semantic 3D representations from unposed images, while AnySplat~\cite{jiang2025anysplat} learns feed-forward Gaussian reconstruction from unconstrained views. UniForward~\cite{tian2025uniforward}, FLEG~\cite{tian2025fleg}, and Uni3R~\cite{sun2025uni3r} further extend this direction by coupling feed-forward Gaussian reconstruction with semantic or language-aligned scene fields. In parallel, semantic Gaussian methods~\cite{guo2024semantic, zhou2024feature3dgs, shi2024language, wang2025language, sheng2025spatialsplat} attach distilled visual-language features, learned feature fields, or language embeddings to 3D Gaussian primitives for open-vocabulary querying and segmentation. These studies are highly relevant to our setting because they reduce the dependence on calibrated cameras or enable semantic reasoning over Gaussian representations. However, they still primarily supervise renderable surfaces and semantic fields through image-space objectives, and they do not explicitly couple the Gaussian representation with a dense semantic occupancy field during training. In contrast, COVScene derives semantic occupancy from pose-free Gaussian primitives through differentiable volumetric lifting, so that rendering, semantic querying, and physical space reasoning are optimized within a single coupled representation.\looseness=-1

\vspace{-10pt}
\subsection{Semantic Occupancy Prediction}
Semantic occupancy prediction provides a dense representation for reasoning about both scene geometry and semantic layout. Large-scale benchmarks, including MonoScene~\cite{cao2022monoscene}, SurroundOcc~\cite{wei2023surroundocc}, Occ3D~\cite{tian2023occ3d}, OpenOccupancy~\cite{wang2023openoccupancy}, SemanticKITTI~\cite{behley2019semantickitti} and Nuplan-Occ~\cite{li2026scaling}, have promoted the development of robust occupancy models through standardized data and evaluation protocols. Fully supervised outdoor methods define several influential design choices~\cite{
li2024one,huang2023tpvformer,zhang2023occformer}. Transformer-based models such as TPVFormer~\cite{huang2023tpvformer}, HTCL~\cite{li2024hierarchical},  VoxFormer~\cite{li2023voxformer}, and OccFormer~\cite{zhang2023occformer} lift image features into dense or semi-dense 3D representations, including voxel grids and tri-plane features. To reduce the redundancy of dense voxels, GaussianFormer~\cite{huang2024gaussianformer} and GaussianFormer-2~\cite{huang2024gaussianformer2} represent occupancy with sparse 3D Gaussians. CausalOcc~\cite{chen2025casualocc} further improves semantic consistency by constraining gradient flow in modular lifting pipelines. Because exhaustive 3D labels are expensive, self-supervised and rendering-based alternatives such as SelfOcc~\cite{huang2024selfocc}, GaussianOcc~\cite{gan2024gaussianocc}, SimpleOcc~\cite{gan2024simpleocc}, and OccNeRF~\cite{zhang2025occnerf} reduce the reliance on dense annotations. Indoor semantic occupancy remains less explored. MonoScene~\cite{cao2022monoscene} and ISO~\cite{yu2024iso} infer dense geometry and semantics from limited monocular observations, while EmbodiedOcc~\cite{wu2025embodiedocc}, EmbodiedOcc$++$~\cite{wang2025embodiedocc_pp}, and SliceOcc~\cite{li2025sliceocc} address occlusion, active perception, and efficient volumetric processing. However, these methods are usually built on voxel-centric formulations and often require stronger supervision or posed inputs. COVScene differs by deriving semantic occupancy directly from pose-free Gaussian primitives through differentiable volumetric lifting, which enables volumetric regularization without introducing a detached occupancy branch.

\section{Methodology}
\label{sec:method}

\begin{figure*}[!t]
    \centering
    \includegraphics[width=0.99\linewidth]{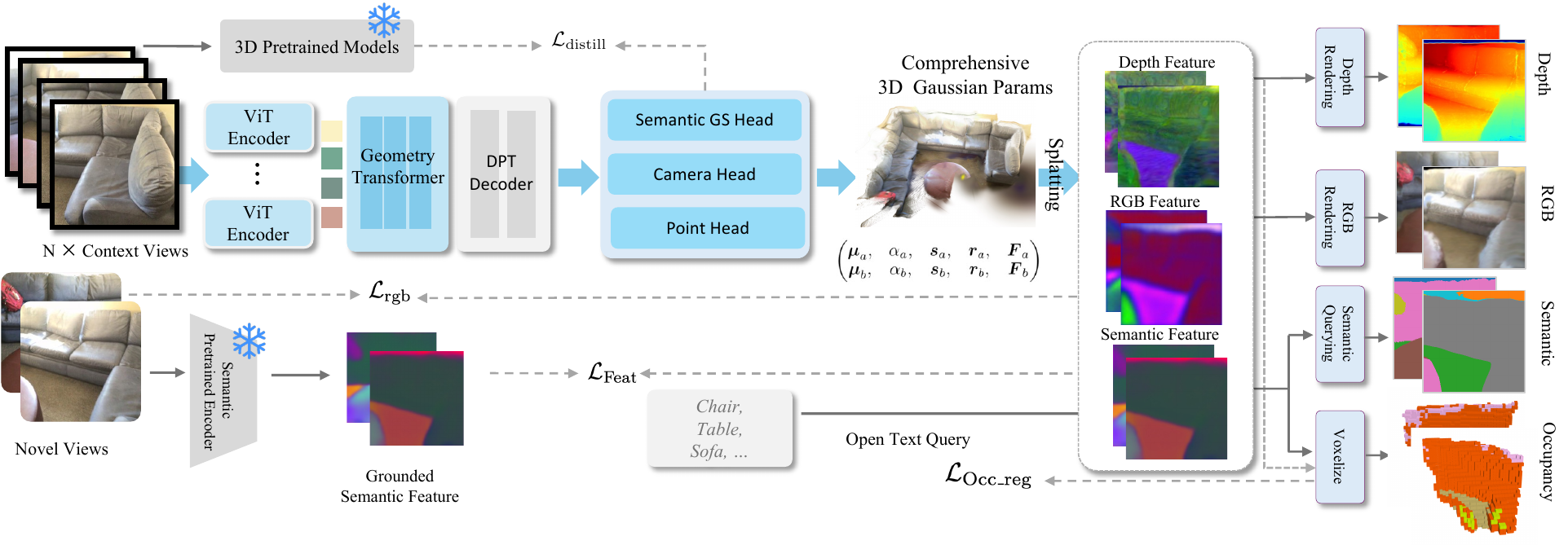}
    \vspace{-10pt}
    
    \caption{Overview of COVScene. Given multi-view unposed images, a semantic-aware Geometry Transformer estimates camera geometry, depth, and semantic Gaussian attributes, while a pretrained 2D semantic feature extractor provides language-aligned feature targets. The resulting semantic 3D Gaussian primitives are differentiably lifted into a dense occupancy field, and voxel semantics are decoded from the lifted feature field at inference time. The shared representation supports novel view synthesis, open-vocabulary segmentation, and semantic occupancy prediction.}
    \label{fig: pipeline}
\end{figure*}

\subsection{Problem Setup}
Given a set of $N$ uncalibrated and unposed RGB images $\mathcal{I} = \{I_i\}_{i=1}^N$, where each $I_i \in \mathbb{R}^{H \times W \times 3}$ captures a sparse observation of a 3D scene, our goal is to recover a globally consistent 3D representation for novel view synthesis, open-vocabulary semantic querying, and semantic occupancy prediction without relying on external Structure-from-Motion (SfM) pipelines. Recent feed-forward 3D Gaussian Splatting (3DGS) methods, such as AnySplat~\cite{jiang2025anysplat}, LSM~\cite{fan2024lsm}, Uni3R~\cite{sun2025uni3r} and MVSplat~\cite{chen2024mvsplat}, have shown that multi-view image features can be efficiently converted into renderable Gaussian primitives. However, these methods mainly operate on \textit{discrete, surface-level} primitives and supervise them through image-space rendering or semantic losses. Even when sparse 2D geometric priors are used, the representation lacks explicit \textit{continuous 3D volumetric constraints} for weakly observed regions. As a result, sparse and unposed inputs can still lead to semi-transparent ``floaters'', hollow structures, or physically implausible occlusion geometry.\looseness=-1

To address this limitation, we propose a \textbf{Volume-Regularized 3DGS} framework that explicitly couples renderable Gaussian primitives with a dense semantic occupancy field during training. The key design is to treat occupancy not as an auxiliary prediction head or a detached post-processing result, but as a differentiable volume lifted from the predicted Gaussian representation. This closed-loop Gaussian-volume coupling allows volumetric losses to update the same opacity, geometry, covariance, and semantic features that are used for novel view synthesis and open-vocabulary segmentation. Therefore, the lifted occupancy field acts as a dense physical prior over free and occupied space, while the underlying Gaussian representation retains the rendering efficiency and semantic flexibility of 3DGS.
Formally, as shown in Fig.~\ref{fig: pipeline}, COVScene first predicts semantic Gaussian primitives from unposed input images and then derives semantic occupancy through differentiable volumetric lifting:
\begin{equation}
\mathcal{G} = F_\theta(\mathcal{I}), \quad
\mathcal{O} = \mathcal{L}_{\mathrm{lift}}(\mathcal{G}).
\end{equation}
Here, $\mathcal{G} = { (\mu_g, \sigma_g, r_g, s_g, c_g, f_g) }{g=1}^G$ denotes the predicted 3D Gaussian set; $\mu_g$, $\sigma_g$, $r_g$, $s_g$, and $c_g$ represent the center, opacity, rotation, scale, and color of primitive $g$, respectively. $f_g \in \mathbb{R}^d$ encodes open-vocabulary semantic features aligned with 2D vision-language models. The dense semantic occupancy field $\mathcal{O}$ is derived from $\mathcal{G}$ alone through $\mathcal{L}_{\mathrm{lift}}$; thus, volumetric regularization on $\mathcal{O}$ propagates gradients back to $\mathcal{G}$ rather than being confined to a separate occupancy branch. This makes the lifted occupancy a native training-time representation of the unified model and distinguishes COVScene from post-processed Gaussian voxelization.\looseness=-1

\vspace{-5pt}
\subsection{Geometry Encoding and Multi-Task Decoding}
Recovering a consistent Gaussian representation from uncalibrated views requires the model to infer camera geometry and scene structure directly from images. To avoid dependence on external Structure-from-Motion pipelines, COVScene adopts a semantic-aware Geometry Transformer as the shared feature encoder. Its geometry-oriented attention blocks are initialized from the pretrained VGGT model~\cite{wang2025vggt}, while additional semantic fusion layers and task-specific prediction heads are introduced to encode language-aligned features and Gaussian attributes. The encoder is designed to capture both local visual evidence within each view and cross-view correspondences across the unposed image set, so that downstream prediction heads can estimate camera parameters, depth, and semantic Gaussian attributes in a unified feed-forward pass.

Given the unposed input images $\mathcal{I} = \{I_i\}_{i=1}^{N}$, the Geometry Transformer maps them into multi-view latent tokens:
\begin{equation}
\{Z_i\}_{i=1}^{N} = \Phi_{\mathrm{geo}}(\mathcal{I}),
\end{equation}
where each $Z_i$ contains dense image tokens together with global tokens for camera and scene-level reasoning. The encoder first embeds image patches and then alternates between intra-view self-attention and inter-view attention. Intra-view attention models local image context, while inter-view attention establishes geometric correspondences across different views. The resulting 3D-aware tokens provide the geometric prior needed for reconstructing scenes from uncalibrated images, without assuming known camera poses or an external SfM reconstruction. \looseness=-1

These latent tokens are decoded by a Dense Prediction Transformer (DPT)-based decoder with three specialized prediction heads. The \textbf{Camera Head} decodes the global camera tokens to predict relative camera extrinsics and intrinsics. The \textbf{Depth Head} estimates per-pixel depth maps from the dense image tokens; combining the predicted depths with the inferred camera parameters gives the 3D centers $\{\mu_g\}$ of the Gaussian primitives through unprojection. The \textbf{Semantic Gaussian Head} fuses Transformer features with high-resolution features from a pretrained 2D semantic encoder to predict opacity $\sigma_g$, rotation $r_g$, scale $s_g$, color $c_g$, and the open-vocabulary semantic feature $f_g$ for each primitive. Because $f_g$ is attached to the same 3D Gaussian feature field used for rendering, source and target semantic maps are rendered from a shared representation, which improves cross-view semantic consistency and supports open-vocabulary querying.

\subsection{Differentiable Volumetric Lifting for Semantic Occupancy}

The differentiable volumetric lifting module converts the predicted Gaussian set into a dense semantic occupancy field inside the training computation graph. Standard 3DGS represents a scene using discrete surface primitives, which are efficient to render but do not explicitly model empty space. This limitation is problematic for sparse and unposed inputs because ambiguous density can be assigned to weakly observed regions while still satisfying image-space supervision. To introduce volumetric reasoning, we define a voxelized 3D domain $V \subset \Omega \subset \mathbb{R}^{3}$ and evaluate, at each voxel center $\mathbf{x} \in V$, an occupancy probability $O(\mathbf{x}) \in [0,1]$ together with a semantic feature $F(\mathbf{x}) \in \mathbb{R}^{d}$.

Given the Gaussian set $\mathcal{G}$, each primitive contributes a local density to spatial position $\mathbf{x}$. For a Gaussian primitive $g$, this density is computed as
\begin{equation}
    \tau_g(\mathbf{x}) = \sigma_g \cdot \exp\Big(-\frac{1}{2} (\mathbf{x} - \mu_g)^\top \Sigma_g^{-1} (\mathbf{x} - \mu_g)\Big),
\end{equation}
where $\Sigma_g$ is the covariance matrix parameterized by scale $s_g$ and rotation $r_g$. We define $\mathcal{N}(\mathbf{x})$ as the local support of $\mathbf{x}$, containing the Gaussians that fall within a truncation radius of the voxel center, with a top-$k$ cap when more candidates are present. Gaussians outside this local support have negligible density and are ignored for efficiency. The voxel-size factor and density scale are absorbed into $\tau_g(\mathbf{x})$. The occupancy probability is then obtained by accumulating the densities from $\mathcal{N}(\mathbf{x})$ using the standard volumetric transmittance form:

\begin{equation}
O(\mathbf{x}) = 1 - \exp\Big(-\sum_{g \in \mathcal{N}(\mathbf{x})} \tau_g(\mathbf{x})\Big).
\end{equation}
The corresponding voxel-wise semantic feature is computed by density-weighted aggregation of the Gaussian semantic features:
\begin{equation}
F(\mathbf{x}) =
\frac{\sum_{g \in \mathcal{N}(\mathbf{x})} \tau_g(\mathbf{x}) f_g}
{\sum_{g \in \mathcal{N}(\mathbf{x})} \tau_g(\mathbf{x}) + \epsilon},
\end{equation}
where $\epsilon$ is a small constant for numerical stability. This construction maps $\mathcal{G}$ to a dense field $(O,F)$ without introducing an auxiliary occupancy head. At inference time, semantic occupancy is decoded from the same voxel feature field. Given a text or class vocabulary $\mathcal{C}$ with normalized embeddings $\{e_c\}_{c \in \mathcal{C}}$ in the same semantic space, each occupied voxel is assigned an open-vocabulary semantic label by
\begin{equation}
\hat{s}(\mathbf{x}) = \arg\max_{c \in \mathcal{C}} \cos\big(F(\mathbf{x}), e_c\big),
\quad \mathrm{for}\ O(\mathbf{x}) > \eta,
\end{equation}
where $\eta$ is the occupancy threshold, and voxels with $O(\mathbf{x}) \leq \eta$ are treated as free space.

Because the lifting operation is fully differentiable, any volumetric regularization applied to the dense field propagates gradients back to the same Gaussian opacity, centers, covariance, and semantic features used for rendering and open-vocabulary segmentation. This property distinguishes COVScene from evaluation-only post-voxelization, in which trained Gaussians are converted to voxels after optimization, and no volumetric gradient is used during training. In COVScene, the lifted occupancy field is therefore a training-time geometric constraint on the Gaussian representation, while voxel semantics are obtained from the language-aligned feature field at inference time rather than from direct voxel-level semantic supervision.

\subsection{Physics-Informed Optimization via Entropy Regularization}

Optimizing COVScene requires supervision that can constrain unposed 3D reconstruction without explicit 3D voxel labels. Image-space rendering losses provide strong appearance supervision, but they do not sufficiently constrain geometry in occluded or weakly observed regions. We therefore combine 2D rendering supervision, feature-level semantic alignment, geometric foundation distillation, and occupancy entropy regularization into a single end-to-end objective:
\begin{equation}
\begin{aligned}
\mathcal{L}_{\mathrm{total}} =
&\lambda_{\mathrm{photo}} \mathcal{L}_{\mathrm{photo}}
+ \lambda_{\mathrm{sem}} \mathcal{L}_{\mathrm{sem}} \\
&+ \lambda_{\mathrm{distill}} \mathcal{L}_{\mathrm{distill}}
+ \lambda_{\mathrm{occ\_reg}} \mathcal{L}_{\mathrm{occ\_reg}}.
\end{aligned}
\end{equation}
where the balancing weights are empirically set to $\lambda_{\mathrm{photo}} = 1.0$, $\lambda_{\mathrm{sem}} = 0.3$, $\lambda_{\mathrm{distill}} = 0.7$, and $\lambda_{\mathrm{occ\_reg}} = 0.25$.

\noindent\textbf{Geometric Foundation Distillation ($\mathcal{L}_{\mathrm{distill}}$)} 
To bootstrap geometry in the strictly unposed setting, we distill camera and spatial cues from a frozen geometry foundation model, such as VGGT~\cite{wang2025vggt}. The frozen teacher is used only to generate pseudo targets, while the trainable COVScene encoder includes additional semantic fusion layers and task-specific heads. Let $\Pi$, $D$, and $P$ denote the predicted camera parameters, depth maps, and 3D point maps, and let $\tilde{\Pi}$, $\tilde{D}$, and $\tilde{P}$ denote the corresponding pseudo targets from the frozen teacher. The distillation loss is defined as
\begin{equation}
\mathcal{L}_{\mathrm{distill}} =
\gamma_{\mathrm{pose}} \|\Pi - \tilde{\Pi}\|_1
+ \gamma_{\mathrm{depth}} \|D - \tilde{D}\|_1
+ \gamma_{\mathrm{point}} \|P - \tilde{P}\|_1,
\end{equation}
where $\gamma_{\mathrm{pose}}$, $\gamma_{\mathrm{depth}}$, and $\gamma_{\mathrm{point}}$ balance the pose, depth, and point-map terms. This guidance provides an initial multi-view geometric scaffold, accelerates convergence, and reduces scale ambiguity without relying on SfM point clouds.\looseness=-1

\noindent\textbf{Photometric and Semantic Supervision ($\mathcal{L}_{\mathrm{photo}}$, $\mathcal{L}_{\mathrm{sem}}$)} 
We differentially splat the predicted Gaussians onto target views. The photometric loss $\mathcal{L}_{\mathrm{photo}}$ combines the $\mathcal{L}_1$ and LPIPS distances between rendered and ground-truth RGB images. For semantic supervision, we render the Gaussian semantic features $f_g$ into 2D feature maps and align them with pseudo target features from a frozen vision-language model, such as CLIP~\cite{clip}, using a cosine similarity loss $\mathcal{L}_{\mathrm{sem}}$. This feature-level supervision avoids fixed-category pseudo-labels and preserves open-vocabulary querying via text-feature matching.\looseness=-1

\begin{figure}[!t]
  \centering
  \includegraphics[width=0.95\linewidth]{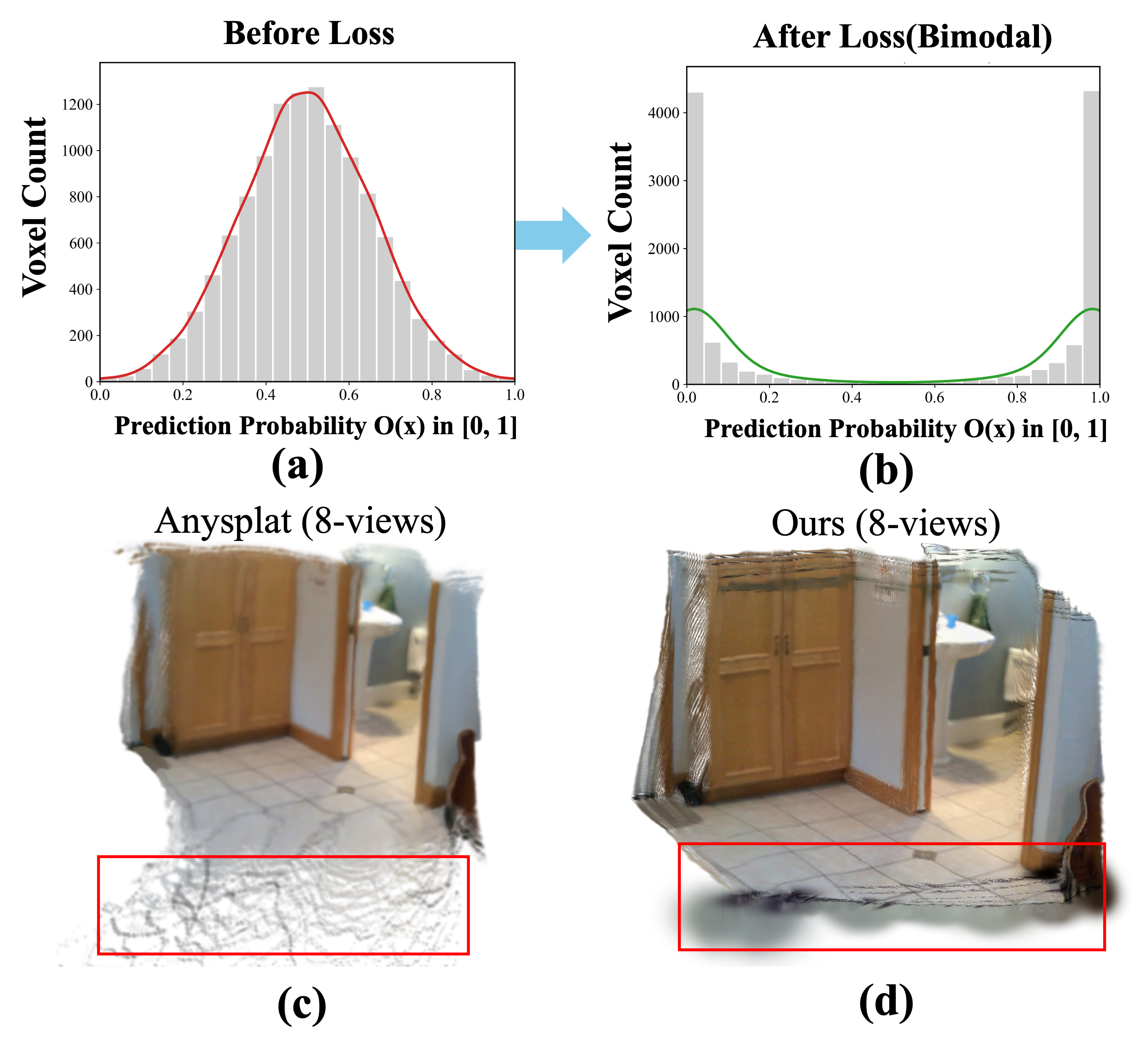}
  \vspace{-10pt}  
  \caption{\textbf{Effect of Occupancy Entropy Regularization ($\mathcal{L}_{\mathrm{occ\_reg}}$).} \textbf{Left:} Without 
$\mathcal{L}_{\mathrm{occ\_reg}}$, unposed 3DGS produces floating density artifacts in empty space. The proposed regularization suppresses these artifacts. \textbf{Right:} The entropy loss penalizes ambiguous occupancy probabilities \mbox{($O(\mathbf{x}) \approx 0.5$)} and encourages a bimodal physical state, where each region is either free or occupied.}
  \vspace{-15pt}  
\label{fig: occ_regulation}
\end{figure}

\noindent\textbf{Occupancy Entropy Regularization ($\mathcal{L}_{\mathrm{occ\_reg}}$)} 
Although rendering supervision and foundation distillation provide the main geometric constraints, regions with limited observations can still yield ambiguous occupancy values in the lifted field. The entropy term is therefore not used as direct voxel supervision; instead, it regularizes the occupancy field induced by the learned Gaussian geometry. We impose a bimodal prior on the lifted field so that each spatial location tends to be either free, with $O(\mathbf{x}) \to 0$, or occupied, with $O(\mathbf{x}) \to 1$. We implement this prior by minimizing the entropy of the predicted occupancy field:
\begin{equation}
\begin{split}
\mathcal{L}_{\mathrm{occ\_reg}} = - \frac{1}{|V|} \sum_{\mathbf{x} \in V}
\big[ O(\mathbf{x}) \log \big(O(\mathbf{x}) + \epsilon\big) \\
{}+ \big(1 - O(\mathbf{x})\big)
\log \big(1 - O(\mathbf{x}) + \epsilon\big) \big],
\end{split}
\end{equation}
where $\epsilon$ denotes a small constant for numerical stability. The entropy reaches its maximum when $O(\mathbf{x}) \approx 0.5$ and decreases as the prediction approaches either 0 or 1. Therefore, conditioned on the Gaussian field learned from rendering, semantic alignment, and geometric distillation, $\mathcal{L}_{\mathrm{occ\_reg}}$ penalizes residual uncertain density and sharpens the free/occupied decision, as shown in Fig.~\ref{fig: occ_regulation}. Since $O(\mathbf{x})$ is lifted from the Gaussian representation, the entropy gradients are propagated through $\mathcal{L}_{\mathrm{lift}}$ to the corresponding Gaussian opacity and geometry parameters. This training-time volumetric feedback reduces ambiguous floating density and differs from post-voxelization baselines, where occupancy is computed only after Gaussian optimization.\looseness=-1

\section{Experiments}

\begin{table*}[!t]
\centering
\setlength{\tabcolsep}{3pt}
\renewcommand{\arraystretch}{1.1}
\caption{Comparative results for multi-task 3D reconstruction and semantic segmentation. Source views are used as context inputs, while target views are disjoint held-out views used only for evaluation. Bold and underlined values indicate the best and second-best performance, respectively. COVScene provides competitive rendering quality while improving open-vocabulary segmentation in the unified feed-forward setting. LSM~\cite{fan2024lsm}\bh{$^{*}$} is post-processed to support multi-view evaluation.}

\resizebox{\textwidth}{!}{ 
\begin{tabular}{l | cc ccccc | cc ccccc}
\toprule
\multirow{3}{*}{Method} & \multicolumn{7}{c|}{2 views} & \multicolumn{7}{c}{8 views} \\
\cmidrule(lr){2-8} \cmidrule(lr){9-15}
& \multicolumn{2}{c}{Source view} & \multicolumn{5}{c|}{Target view} & \multicolumn{2}{c}{Source view} & \multicolumn{5}{c}{Target view} \\
\cmidrule(lr){2-3} \cmidrule(lr){4-8} \cmidrule(lr){9-10} \cmidrule(lr){11-15}
& mIoU$\uparrow$ & Acc$\uparrow$ & mIoU$\uparrow$ & Acc$\uparrow$ & PSNR$\uparrow$ & SSIM$\uparrow$ & LPIPS$\downarrow$ & mIoU$\uparrow$ & Acc$\uparrow$ & mIoU$\uparrow$ & Acc$\uparrow$ & PSNR$\uparrow$ & SSIM$\uparrow$ & LPIPS$\downarrow$ \\
\midrule
AnySplat~\cite{jiang2025anysplat}        & - & - & - & - & 22.22 & \underline{0.774} & 0.293 & - & - & - & - & \textbf{23.25} & \textbf{0.786} & 0.275 \\
LSeg~\cite{Li2022Lseg}       & 52.40 & 77.30 & 51.20 & 78.40 & - & - & - & 53.20 & 77.80 & 53.30 & \textbf{78.50} & - & - & - \\
LSM~\cite{fan2024lsm}\bh{$^{*}$}        & 51.40 & 77.10 & 51.00 & 76.60 & \underline{22.58} & 0.758 & \textbf{0.252} & 53.40 & 77.90 & 51.90 & 76.60 & 20.58 & 0.742 & 0.435 \\
Feature-3DGS~\cite{zhou2024feature3dgs}    & 44.50 & 73.00 & 43.20 & 71.20 & 20.67 & 0.617 & 0.382 & 45.40 & 73.40 & 44.20 & 72.10 & 20.89 & 0.631 & 0.375 \\
\rowcolor{blue!10} Ours & \textbf{53.90} & \textbf{78.10} & \textbf{53.20} & \textbf{79.20} & \textbf{23.13} & \textbf{0.782} & \underline{0.265} & \textbf{54.90} & \textbf{78.50} & \textbf{55.20} & \underline{78.30} & \underline{23.15} & \underline{0.784} & \textbf{0.262} \\
\midrule
\midrule
\multirow{3}{*}{Method} & \multicolumn{7}{c|}{16 views} & \multicolumn{7}{c}{32 views} \\
\cmidrule(lr){2-8} \cmidrule(lr){9-15}
& \multicolumn{2}{c}{Source view} & \multicolumn{5}{c|}{Target view} & \multicolumn{2}{c}{Source view} & \multicolumn{5}{c}{Target view} \\
\cmidrule(lr){2-3} \cmidrule(lr){4-8} \cmidrule(lr){9-10} \cmidrule(lr){11-15}
& mIoU$\uparrow$ & Acc$\uparrow$ & mIoU$\uparrow$ & Acc$\uparrow$ & PSNR$\uparrow$ & SSIM$\uparrow$ & LPIPS$\downarrow$ & mIoU$\uparrow$ & Acc$\uparrow$ & mIoU$\uparrow$ & Acc$\uparrow$ & PSNR$\uparrow$ & SSIM$\uparrow$ & LPIPS$\downarrow$ \\
\midrule
AnySplat~\cite{jiang2025anysplat}        & - & - & - & - & \textbf{23.72} & \textbf{0.792} & \underline{0.272} & - & - & - & - & \underline{23.93} & \underline{0.796} & \underline{0.259} \\
LSeg~\cite{Li2022Lseg}       & \underline{54.30} & 78.10 & \underline{53.50} & 77.90 & - & - & - & \underline{54.10} & 77.90 & \underline{53.90} & \underline{78.10} & - & - & - \\
LSM~\cite{fan2024lsm}\bh{$^{*}$}         & 50.20 & 75.60 & 49.30 & 71.20 & 19.53 & 0.672 & 0.478 & 49.50 & 73.20 & 50.10 & 74.10 & 17.72 & 0.614 & 0.531 \\
Feature-3DGS~\cite{zhou2024feature3dgs}    & 47.20 & 75.30 & 46.20 & 74.10 & 21.02 & 0.644 & 0.321 & 47.20 & 75.30 & 47.20 & 75.30 & 22.02 & 0.663 & 0.312 \\
\rowcolor{blue!10} Ours & \textbf{55.40} & \textbf{78.90} & \textbf{55.30} & \textbf{78.70} & \underline{23.56} & \underline{0.787} & \textbf{0.271} & \textbf{55.70} & \textbf{79.00} & \textbf{55.90} & \textbf{79.40} & \textbf{23.95} & \textbf{0.799} & \textbf{0.252} \\
\bottomrule
\end{tabular}
}
  \vspace{-15pt}  
\label{tab: method_comparison}
\end{table*}

\textbf{Implementation Details} 
We evaluate COVScene on three downstream tasks: novel view synthesis (NVS), 3D open-vocabulary segmentation (OVS), and semantic occupancy prediction. Following the data filtering protocol of LSM~\cite{fan2024lsm}, we train the model on ScanNet~\cite{scannet} and ScanNet++~\cite{scannetpp}, yielding approximately 1,500 valid training scenes, and evaluate it on 40 unseen ScanNet scenes with additional validation on ScanNet++. Source views are used as context inputs, while target views are disjoint held-out views used only for evaluation and are not used for training, pseudo-label generation, or model input. For occupancy evaluation, semantic occupancy annotations are constructed following the ISO protocol~\cite{yu2024iso} when native voxel annotations are unavailable, and dataset categories are used only for evaluation rather than as category-level training labels.

For NVS and OVS, we compare COVScene with representative feed-forward and semantic Gaussian baselines, including AnySplat~\cite{jiang2025anysplat}, LSM~\cite{fan2024lsm}, Feature-3DGS~\cite{zhou2024feature3dgs}, and the 2D open-vocabulary model LSeg~\cite{Li2022Lseg}. Since LSM is originally designed for two-view inputs based on DUSt3R~\cite{Wang2024dust3r}, we extend it to multi-view settings using the global alignment strategy of DUSt3R for fair comparison across different input view counts. For semantic occupancy, we compare with fully supervised methods, including MonoScene~\cite{cao2022monoscene}, ISO~\cite{yu2024iso}, and EmbodiedOcc~\cite{wu2025embodiedocc}, as well as the self-supervised baseline SelfOcc~\cite{huang2024selfocc}. NVS is evaluated by PSNR, SSIM, and LPIPS, OVS by mIoU and mAcc, and semantic occupancy by IoU and mIoU.

The semantic-aware Geometry Transformer uses the same layer count as VGGT~\cite{wang2025vggt}. Geometry-oriented blocks are initialized from VGGT and fine-tuned during training, while the newly introduced semantic fusion layers and task-specific heads are initialized randomly. We train all models with the objective in Sec.~\ref{sec:method}. Additional reproducibility details, including volumetric lifting hyperparameters, grid resolutions, loss weights, and optimization settings, are provided in the supplementary material.\looseness=-1

\subsection{Multi-Task Comparison}

\begin{figure}[!t]
    \centering
    \includegraphics[width=0.95\linewidth]{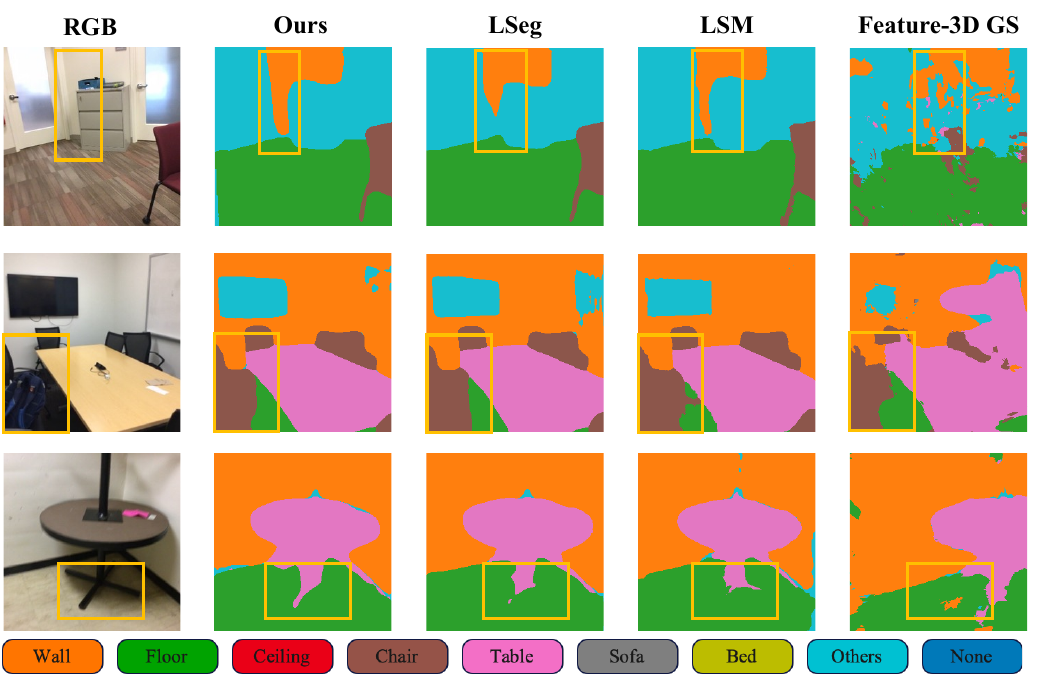}
    \caption{\textbf{Qualitative comparison of open-vocabulary segmentation.} We evaluate segmentation on common ScanNet categories. COVScene produces more consistent 3D-aware semantic maps than feed-forward 3D baselines while remaining competitive with the 2D open-vocabulary model.} 
  \vspace{-15pt}  
    \label{fig: experiment_semantic}
\end{figure}

\noindent\textbf{Open-Vocabulary Segmentation} As shown in Fig.~\ref{fig: experiment_semantic} and Tab.~\ref{tab: method_comparison}, COVScene achieves strong open-vocabulary segmentation performance across both source and target views. Compared with LSeg, which performs 2D open-vocabulary segmentation independently for each view, COVScene produces competitive or better mIoU while maintaining a single 3D semantic Gaussian field. The advantage becomes more visible as the number of input views increases: at $32$ views, COVScene reaches $55.90$ target-view mIoU and $79.40$ target-view accuracy. In contrast, LSM does not benefit consistently from additional views, with source-view mIoU decreasing from $51.40$ under $2$ views to $49.50$ under $32$ views. These results suggest that the shared 3D semantic representation improves cross-view semantic consistency under multi-view inputs. \looseness=-1

\begin{figure}[!t]
    \centering
    \includegraphics[width=0.96\linewidth]{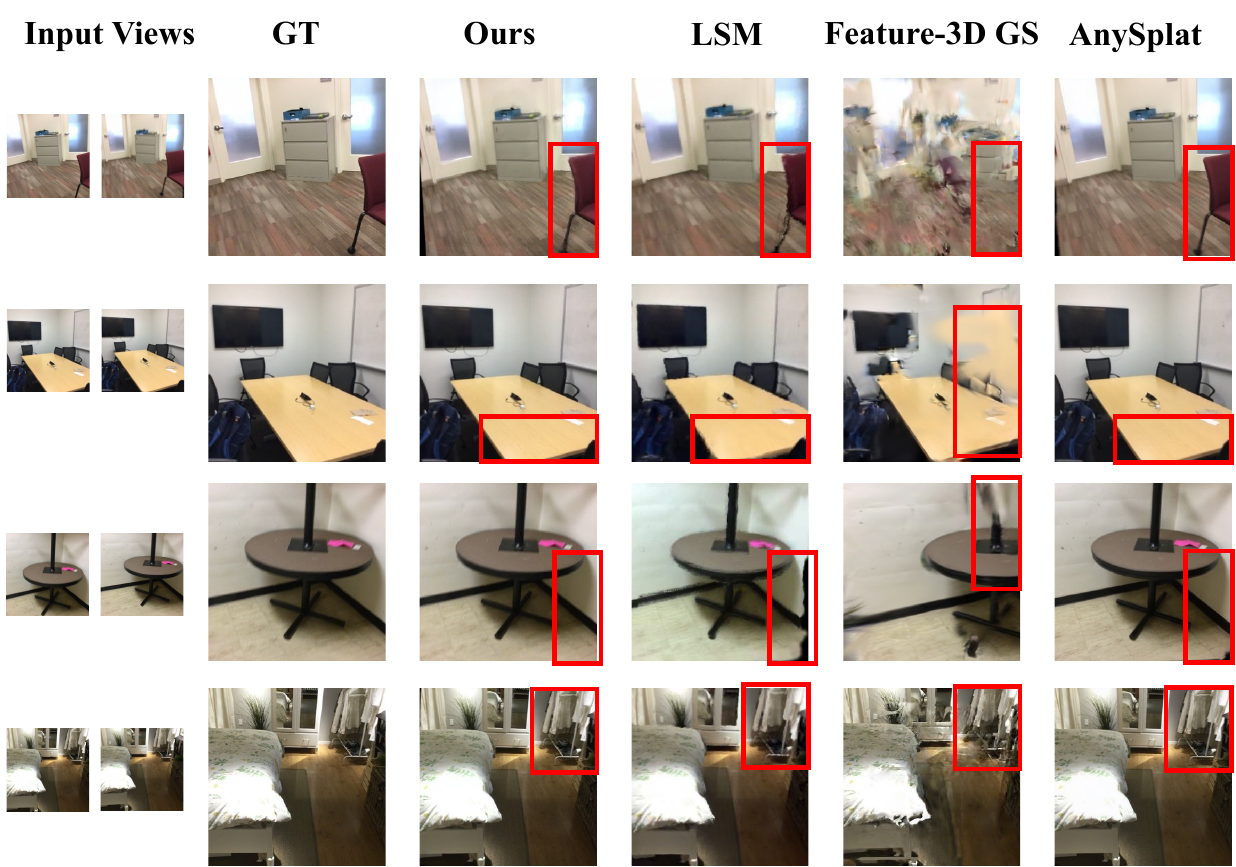}
    \caption{\textbf{Novel view synthesis comparison.} COVScene maintains competitive rendering quality while supporting semantic segmentation and occupancy prediction from the same representation.} 
  \vspace{-15pt}  
    \label{fig: experiment_nvs}
\end{figure}

\noindent\textbf{Novel View Synthesis} As shown in Fig.~\ref{fig: experiment_nvs} and Tab.~\ref{tab: method_comparison}, COVScene preserves competitive rendering quality while adding semantic and occupancy reasoning to the Gaussian representation. Under the $2$-view setting, COVScene obtains the best PSNR and SSIM among the compared methods, with slightly higher LPIPS than LSM. Under $8$ and $16$ views, AnySplat remains stronger in PSNR and SSIM because it is optimized specifically for NVS, whereas COVScene obtains the best LPIPS. With $32$ input views, COVScene achieves the best PSNR, SSIM, and LPIPS among all baselines. These results indicate that the proposed representation does not trade off rendering quality for semantic and occupancy prediction, and that the unified model can effectively use additional views. \looseness=-1

\begin{figure}[!t]
    \centering
    \includegraphics[width=0.96\linewidth]{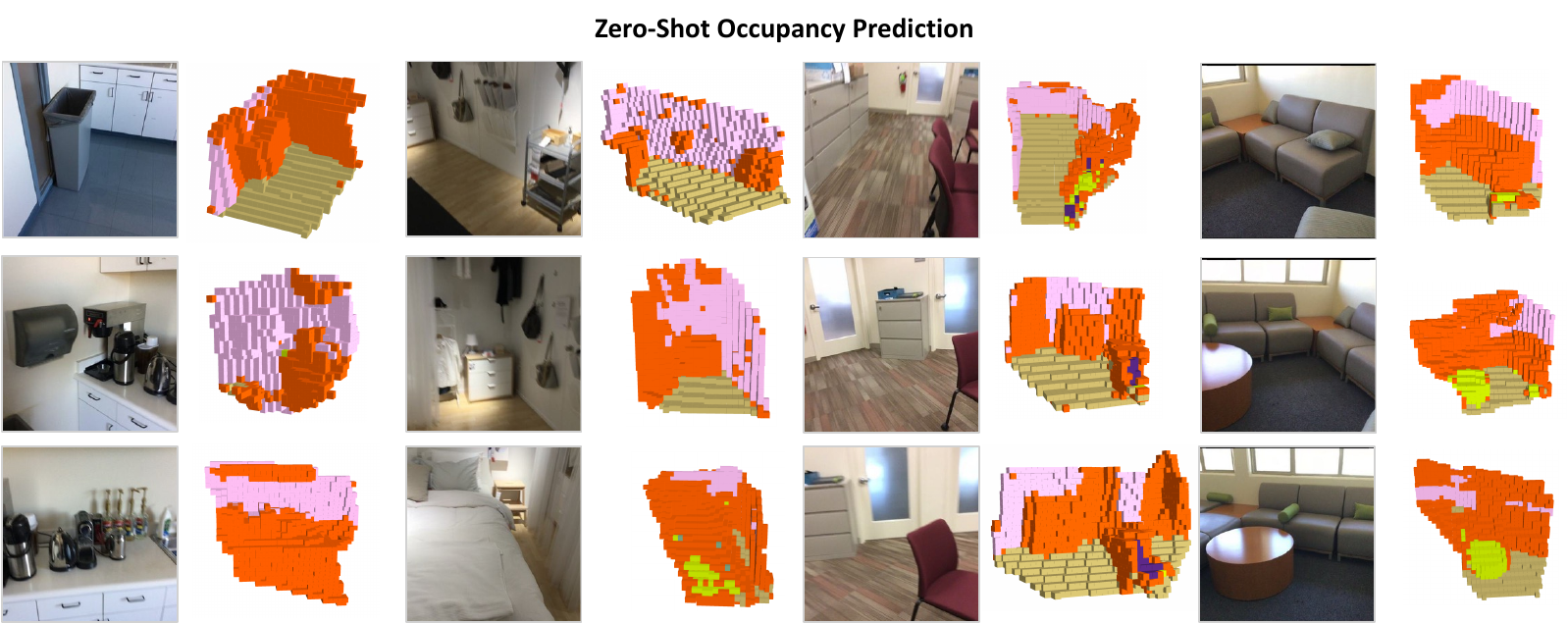}
    \caption{Qualitative results of zero-shot occupancy prediction on in-the-wild images. The visualizations show plausible spatial structures and semantic layouts for scenes without available ground-truth semantic occupancy annotations.} 
      \vspace{-15pt}  
    \label{fig: experiment_occ}
\end{figure}

\noindent\textbf{Semantic Occupancy Prediction} Tab.~\ref{tab: occ_compare} compares COVScene with supervised occupancy methods and self-supervised baselines. A clear gap remains between methods trained with explicit 3D labels and those trained without direct voxel supervision, which reflects the difficulty of semantic occupancy prediction under weak supervision. Within the no-3D-label setting, COVScene improves over SelfOcc from $7.68$ to $18.32$ IoU and from $6.93$ to $17.78$ mIoU. The gains appear across most semantic classes, including structural regions such as walls and windows as well as object categories such as chairs, beds, and sofas. Qualitative zero-shot results in Fig.~\ref{fig: experiment_occ} further illustrate that the learned representation can infer plausible occupancy layouts for images without available 3D ground-truth annotations.\looseness=-1

\begin{table*}[!t]
    \centering
    \setlength{\tabcolsep}{0.008\linewidth}
    \caption{\textbf{Quantitative comparison of semantic occupancy prediction.} The upper block contains methods trained with explicit 3D labels, while the lower block contains methods trained without direct 3D voxel labels. `No 3D labels' indicates that direct voxel-level supervision is not used during training, and bold values mark the best results within this setting.}
    \resizebox{1.0\linewidth}{!}{
        \begin{tabular}{l c c c c c c c c c c c c c c}
            \toprule
            Method & \bh{No 3D labels}
            & {IoU$\uparrow$}
            & \rotatebox{90}{\parbox{1.5cm}{\textcolor{ceiling}{$\blacksquare$} ceiling}} 
            & \rotatebox{90}{\textcolor{floor}{$\blacksquare$} floor}
            & \rotatebox{90}{\textcolor{wall}{$\blacksquare$} wall} 
            & \rotatebox{90}{\textcolor{window}{$\blacksquare$} window} 
            & \rotatebox{90}{\textcolor{chair}{$\blacksquare$} chair} 
            & \rotatebox{90}{\textcolor{bed}{$\blacksquare$} bed} 
            & \rotatebox{90}{\textcolor{sofa}{$\blacksquare$} sofa} 
            & \rotatebox{90}{\textcolor{table}{$\blacksquare$} table} 
            & \rotatebox{90}{\textcolor{tvs}{$\blacksquare$} tvs} 
            & \rotatebox{90}{\textcolor{furniture}{$\blacksquare$} furniture} 
            & \rotatebox{90}{\textcolor{objects}{$\blacksquare$} objects} 
            & mIoU$\uparrow$\\
            \midrule

            MonoScene~\cite{cao2022monoscene}&  \textcolor{red}{\usym{2717}} & 41.60 & 15.17 & 44.71 & 22.41 & 12.55 & 26.11 & 27.03 & 35.91 & 28.32 & 6.57 & 32.16 & 19.84 & 24.62 \\
            ISO~\cite{yu2024iso}&  \textcolor{red}{\usym{2717}} & 42.16 & 19.88 & 41.88 & 22.37 & 16.98 & 29.09 & 42.43 & 42.00 & 29.60 & 10.62 & 36.36 & 24.61 & 28.71 \\ 
            EmbodiedOcc~\cite{wu2025embodiedocc}&  \textcolor{red}{\usym{2717}}  & 53.95 & 40.90 & 50.80 & 41.90 & 33.00 & 41.20 & 55.20 & 61.90 & 43.80 & 35.40 & 53.50 & 42.90 & 45.50 \\ \midrule
            SelfOcc~\cite{huang2024selfocc} & \textcolor{ForestGreen}{\usym{2713}}  & 7.68 & 2.31 & 10.54 & 8.23 & 6.75 & 10.34 & 9.57 & 8.72 & 4.36 & 1.21 & 9.67 & 4.52 & 6.93 \\
            Ours & \textcolor{ForestGreen}{\usym{2713}}  & \textbf{18.32} & \textbf{10.21} & \textbf{22.19} & \textbf{20.78} & \textbf{21.34} & \textbf{23.18} & \textbf{23.41} & \textbf{22.79} & \textbf{13.24} & \textbf{12.42} & \textbf{15.26} & \textbf{10.79} & \textbf{17.78}\\
            \bottomrule
        \end{tabular}
    }
    \label{tab: occ_compare}
    \vspace{-15pt}
\end{table*}

\subsection{Ablation Study}
\label{ablation}

\begin{table}[!t]
  \centering
  \scriptsize
  \caption{\textbf{Controlled comparison with post-voxelization under the 2-view setting.} `Post-voxel.' converts trained semantic Gaussians to occupancy only at evaluation time, without training-time volumetric feedback. $^\dagger$ includes the extra post-voxelization time.}
  \setlength{\tabcolsep}{4pt}
  \resizebox{\linewidth}{!}{
    \begin{tabular}{lcccc}
      \toprule
      Method / Setting & PSNR$\uparrow$ & OVS mIoU$\uparrow$ & Occ. mIoU$\uparrow$ & Time (s)$\downarrow$ \\
      \midrule
      AnySplat~\cite{jiang2025anysplat} & 22.22 & -- & -- & 0.161 \\
      AnySplat+sem.+post-voxel. & 21.79 & 0.389 & 7.32 & 0.545$^\dagger$ \\
      Uni3R~\cite{sun2025uni3r} & 23.21 & 0.523 & -- & 0.172 \\
      Uni3R+post-voxel. & 23.21 & 0.523 & 12.63 & 0.480$^\dagger$ \\
      Ours-GS+post-voxel. & 22.89 & 0.521 & 12.79 & 0.329$^\dagger$ \\
      Ours w/o Lift. & 20.28 & 0.376 & 8.47 & 0.173 \\
      Ours w/o Ent. & 21.21 & 0.491 & 10.34 & 0.165 \\
      \rowcolor{blue!10} COVScene & \textbf{23.13} & \textbf{0.532} & \textbf{17.21} & \textbf{0.159} \\
      \bottomrule
    \end{tabular}
  }
    \vspace{-10pt}
  \label{tab: post_voxel_ablation}
\end{table}

\begin{table}[!t]
  \centering
  \footnotesize
  
  \caption{\textbf{Ablation Study of Key Components on ScanNet dataset.} 
  }
  \setlength{\tabcolsep}{5pt} 
  \resizebox{\linewidth}{!}{
    \begin{tabular}{lcccc} 
      \toprule
      Configuration & PSNR~($\uparrow$) & SSIM~($\uparrow$) & LPIPS~($\downarrow$) & mIoU~($\uparrow$) \\
      \midrule
      \textbf{Full Framework (Ours)} & \textbf{23.15} & \textbf{0.784} & \textbf{0.262} & \textbf{0.552} \\
      \midrule
      w/o Occupancy Entropy Reg. ($\mathcal{L}_{\mathrm{occ\_reg}}$) & 22.51 & 0.750 & 0.285  & 0.518 \\
      w/o Pose Distillation                                          & 12.50 & 0.580 & 0.560  & 0.224 \\
      w/o Differentiable Lifting                                     & 20.34 & 0.672 & 0.331  & 0.376 \\
      \bottomrule
    \end{tabular}
  }
  \label{tab: ablation}
  \vspace{-10pt}
\end{table}

We perform ablation studies to validate the role of training-time volumetric coupling and the proposed components. Tab.~\ref{tab: post_voxel_ablation} compares COVScene with evaluation-only post-voxelization baselines under the 2-view setting. Directly converting trained semantic Gaussians to occupancy improves over methods without occupancy output, but remains clearly weaker than COVScene. Compared with AnySplat+sem.+post-voxel. and Uni3R+post-voxel., COVScene improves Occ. mIoU by $9.89$ and $4.58$, respectively, while avoiding the extra post-voxelization time. This result supports that the occupancy gain comes from training-time Gaussian-volume coupling rather than a detached conversion step.

We further ablate key components of COVScene under the 8-view setting in Tab.~\ref{tab: ablation}.

\noindent\textbf{Impact of Pose Distillation} Removing pose distillation tests its role in unposed reconstruction. Without this guidance, the network lacks geometry constraints to establish initial multi-view correspondences. Consequently, scene geometry collapses, severely degrading novel view synthesis (PSNR decreases to 12.50) and semantic segmentation (mIoU drops to 0.224). This indicates that distilling pose priors from geometric foundation models is essential for optimizing uncalibrated scenes.\looseness=-1

\noindent\textbf{Effectiveness of Differentiable Lifting} This module bridges discrete Gaussians and the continuous occupancy field. Removing it restricts supervision strictly to 2D projections without 3D volumetric constraints. This degrades 3D scene understanding (mIoU drops from 0.552 to 0.376) and heavily impacts rendering quality (PSNR drops from 23.15 to 20.34). This dual degradation proves that mapping Gaussians to a continuous volume allows spatial constraints to back-propagate and regularize the discrete primitives, maintaining a unified architecture.\looseness=-1

\noindent\textbf{Occupancy Entropy Regularization} Finally, we evaluate the bimodal physical prior ($\mathcal{L}_{\mathrm{occ\_reg}}$). Removing this loss fills unobserved free space with artifacts, as the lack of penalty allows the network to satisfy 2D photometric losses using imprecise geometries. Omitting this regularization decreases rendering PSNR by 0.64 and segmentation mIoU by $3.4\%$. This confirms that enforcing structural certainty successfully suppresses floaters and sharpens semantic boundaries.

\section{Conclusion and Future Work}

In this paper, we presented COVScene, a pose-free feed-forward framework for comprehensive 3D scene understanding from sparse, unposed images. The central idea is to couple renderable semantic Gaussians with a dense occupancy field through differentiable volumetric lifting, so that volumetric regularization provides training-time feedback to Gaussian opacity, geometry, and semantic features rather than acting as evaluation-only voxelization. This coupled representation supports novel view synthesis, open-vocabulary semantic querying, and semantic occupancy prediction without direct voxel-level supervision. Experiments on ScanNet and ScanNet++ show that COVScene maintains competitive rendering quality, improves open-vocabulary segmentation, and achieves stronger no-3D-label semantic occupancy prediction than the self-supervised baseline. The ablation results further indicate that geometric distillation and differentiable volumetric lifting are important for stable unposed reconstruction and occupancy-aware scene modeling.\looseness=-1

\noindent\textbf{Limitations and Future Work.}
The current framework is evaluated mainly on static indoor scenes, and jointly optimizing rendering, semantic alignment, geometric distillation, and volumetric regularization remains computationally demanding. Future work will focus on improving inference and training efficiency, extending the volumetric coupling to dynamic 4D scenes, and studying online deployment in embodied perception systems.\looseness=-1

\bibliographystyle{IEEEtran}
\bibliography{main}

@String(CVPR= {IEEE Conf. Comput. Vis. Pattern Recog.})

@String(ICCV= {Int. Conf. Comput. Vis.})

@String(ECCV= {Eur. Conf. Comput. Vis.})

@String(TOG= {ACM Trans. Graph.})

@String(ICLR = {Int. Conf. Learn. Represent.})

@String(AAAI = {AAAI})

@String(CVPR  = {CVPR})

@String(ICCV  = {ICCV})

@String(ECCV  = {ECCV})

@String(TOG   = {ACM TOG})

@String(ICLR  = {ICLR})

@article{jiang2025anysplat,
  title={Anysplat: Feed-forward 3d gaussian splatting from unconstrained views},
  author={Jiang, Lihan and Mao, Yucheng and Xu, Linning and Lu, Tao and Ren, Kerui and Jin, Yichen and Xu, Xudong and Yu, Mulin and Pang, Jiangmiao and Zhao, Feng and others},
  journal={ACM Transactions on Graphics (TOG)},
  year={2025},
}

@inproceedings{zhuh2022fusing,
  AUTHOR = {Hu Zhu and Yao, Chen and Zhu, Zheng and Liu, Zhengtao and Jia, Zhenzhong},
  TITLE = {Fusing Panoptic Segmentation and Geometry Information for Robust Visual SLAM in Dynamic Environments},
  BOOKTITLE = {CASE},
  YEAR = {2022},
}

@inproceedings{schonberger2016sfm,
  title={Structure-from-motion revisited},
  author={Schonberger, Johannes L and Frahm, Jan-Michael},
  booktitle={Proceedings of the IEEE conference on computer vision and pattern recognition},
  year={2016}
}

@inproceedings{Wang2024dust3r,
  author = {Wang, Shuzhe and Leroy, Vincent and Cabon, Yohann and Chidlovskii, Boris and Revaud, Jerome},
  booktitle = {Proceedings of the IEEE/CVF Conference on Computer Vision and Pattern Recognition (CVPR)},
  title = {DUSt3R: Geometric 3D Vision Made Easy},
  year = {2024}
}

@inproceedings{wang2025vggt,
  author = {Wang, Jianyuan and Chen, Minghao and Karaev, Nikita and Vedaldi, Andrea and Rupprecht, Christian and Novotny, David},
  booktitle = {Proceedings of the IEEE/CVF Conference on Computer Vision and Pattern Recognition},
  title = {VGGT: Visual Geometry Grounded Transformer},
  year = {2025}}

@Inproceedings{Li2022Lseg,
  author = {Boyi Li and Kilian Q. Weinberger and Serge J. Belongie and V. Koltun and René Ranftl},
  booktitle = {International Conference on Learning Representations},
  title = {Language-driven Semantic Segmentation},
  year = {2022}
}

@article{mueller2022instantngp,
  author = {Thomas M\"uller and Alex Evans and Christoph Schied and Alexander Keller},
  title = {Instant Neural Graphics Primitives with a Multiresolution Hash Encoding},
  journal = {ACM Trans. Graph.},
  year = {2022},
}

@inproceedings{yu2020pixelnerf,
  title={{pixelNeRF}: Neural Radiance Fields from One or Few Images},
  author={Alex Yu and Vickie Ye and Matthew Tancik and Angjoo Kanazawa},
  year={2021},
  booktitle={CVPR},
}

@inproceedings{kplanes_2023,
  title={K-Planes: Explicit Radiance Fields in Space, Time, and Appearance},
  author={Sara Fridovich-Keil and Giacomo Meanti and Frederik Rahbæk Warburg and Benjamin Recht and Angjoo Kanazawa},
  year={2023},
  booktitle={CVPR},
}

@inproceedings{tang2025mvdust3r,
  title={Mv-dust3r+: Single-stage scene reconstruction from sparse views in 2 seconds},
  author={Tang, Zhenggang and Fan, Yuchen and Wang, Dilin and Xu, Hongyu and Ranjan, Rakesh and Schwing, Alexander and Yan, Zhicheng},
  booktitle={Proceedings of the Computer Vision and Pattern Recognition Conference},
  year={2025}
}

@inproceedings{scannet,
  title={Scannet: Richly-annotated 3d reconstructions of indoor scenes},
  author={Dai, Angela and Chang, Angel X and Savva, Manolis and Halber, Maciej and Funkhouser, Thomas and Nie{\ss}ner, Matthias},
  booktitle={Proceedings of the IEEE conference on computer vision and pattern recognition},
  year={2017}
}

@inproceedings{zhou2024feature3dgs,
  title={Feature 3dgs: Supercharging 3d gaussian splatting to enable distilled feature fields},
  author={Zhou, Shijie and Chang, Haoran and Jiang, Sicheng and Fan, Zhiwen and Zhu, Zehao and Xu, Dejia and Chari, Pradyumna and You, Suya and Wang, Zhangyang and Kadambi, Achuta},
  booktitle={Proceedings of the IEEE/CVF Conference on Computer Vision and Pattern Recognition},
  year={2024}
}

@article{guo2024semantic,
  title={Semantic Gaussians: Open-Vocabulary Scene Understanding with 3D Gaussian Splatting},
  author={Jun Guo and Xiaojian Ma and Yue Fan and Huaping Liu and Qing Li},
  journal={arXiv preprint arXiv:2403.15624},
  year={2024},
}

@inproceedings{shi2024language,
  title={Language embedded 3d gaussians for open-vocabulary scene understanding},
  author={Shi, Jin-Chuan and Wang, Miao and Duan, Hao-Bin and Guan, Shao-Hua},
  booktitle={Proceedings of the IEEE/CVF Conference on Computer Vision and Pattern Recognition},
  year={2024}
}

@article{zhang2025roboocc,
  title={RoboOcc: Enhancing the Geometric and Semantic Scene Understanding for Robots},
  author={Zhang Zhang and Qiang Zhang and Wei Cui and Shuai Shi and Yijie Guo and Gang Han and Wen Zhao and Hengle Ren and Renjing Xu and Jian Tang},
  year={2025},
  journal={arXiv preprint arXiv:2504.14604},
}

@article{wang2025language,
  title={Language Embedded 3D Gaussians for Open-Vocabulary Scene Querying},
  author={Wang, Miao and Shi, Jin-Chuan and Guan, Shao-Hua and Duan, Hao-Bin},
  journal={IEEE Transactions on Pattern Analysis and Machine Intelligence},
  year={2025},
}

@inproceedings{qin2024langsplat,
  title={Langsplat: 3d language gaussian splatting},
  author={Qin, Minghan and Li, Wanhua and Zhou, Jiawei and Wang, Haoqian and Pfister, Hanspeter},
  booktitle={Proceedings of the IEEE/CVF Conference on Computer Vision and Pattern Recognition},
  year={2024}
}

@Article{3dgs,
  author       = {Kerbl, Bernhard and Kopanas, Georgios and Leimk{\"u}hler, Thomas and Drettakis, George},
  title        = {3D Gaussian Splatting for Real-Time Radiance Field Rendering},
  journal      = {ACM Transactions on Graphics},
  year         = {2023}
}

@article{wang2024freesplat,
  title={Freesplat: Generalizable 3d gaussian splatting towards free view synthesis of indoor scenes},
  author={Wang, Yunsong and Huang, Tianxin and Chen, Hanlin and Lee, Gim Hee},
  journal={Advances in Neural Information Processing Systems},
  year={2024}
}

@inproceedings{nerf,
  title={NeRF: Representing Scenes as Neural Radiance Fields for View Synthesis},
  author={Ben Mildenhall and Pratul P. Srinivasan and Matthew Tancik and Jonathan T. Barron and Ravi Ramamoorthi and Ren Ng},
  year={2020},
  booktitle={ECCV},
}

@inproceedings{cao2022monoscene,
  title={Monoscene: Monocular 3d semantic scene completion},
  author={Cao, Anh-Quan and De Charette, Raoul},
  booktitle={Proceedings of the IEEE/CVF Conference on Computer Vision and Pattern Recognition},
  year={2022}
}

@inproceedings{kirillov2023sam,
  title={Segment anything},
  author={Kirillov, Alexander and Mintun, Eric and Ravi, Nikhila and Mao, Hanzi and Rolland, Chloe and Gustafson, Laura and Xiao, Tete and Whitehead, Spencer and Berg, Alexander C and Lo, Wan-Yen and others},
  booktitle={Proceedings of the IEEE/CVF International Conference on Computer Vision},
  year={2023}
}

@inproceedings{scannetpp,
  title={Scannet++: A high-fidelity dataset of 3d indoor scenes},
  author={Yeshwanth, Chandan and Liu, Yueh-Cheng and Nie{\ss}ner, Matthias and Dai, Angela},
  booktitle={Proceedings of the IEEE/CVF International Conference on Computer Vision},
  year={2023}
}

@inproceedings{yu2024iso,
  title={Monocular occupancy prediction for scalable indoor scenes},
  author={Yu, Hongxiao and Wang, Yuqi and Chen, Yuntao and Zhang, Zhaoxiang},
  booktitle={European Conference on Computer Vision},
  year={2024},
}

@article{liao2024ovnerf,
  title={OV-NeRF: Open-vocabulary neural radiance fields with vision and language foundation models for 3D semantic understanding},
  author={Liao, Guibiao and Zhou, Kaichen and Bao, Zhenyu and Liu, Kanglin and Li, Qing},
  journal={IEEE Transactions on Circuits and Systems for Video Technology},
  year={2024},
}

@inproceedings{garfield2024,
  author = {Kim, Chung Min and Wu, Mingxuan and Kerr, Justin and Tancik, Matthew and Goldberg, Ken and Kanazawa, Angjoo},
  title = {GARField: Group Anything with Radiance Fields},
  booktitle = {Conference on Computer Vision and Pattern Recognition (CVPR)},
  year = {2024},
}

@article{gao2024maskfield,
  title={Fast and Efficient: Mask Neural Fields for 3D Scene Segmentation},
  author={Gao, Zihan and Li, Lingling and Jiao, Licheng and Liu, Fang and Liu, Xu and Ma, Wenping and Guo, Yuwei and Yang, Shuyuan},
  journal={arXiv preprint arXiv:2407.01220},
  year={2024}
}

@InProceedings{OpenSplat3D,
  author    = {Piekenbrinck, Jens and Schmidt, Christian and Hermans, Alexander and Vaskevicius, Narunas and Linder, Timm and Leibe, Bastian},
  title     = {OpenSplat3D: Open-Vocabulary 3D Instance Segmentation using Gaussian Splatting},
  booktitle = {Proceedings of the Computer Vision and Pattern Recognition Conference (CVPR) Workshops},
  year      = {2025},
}

@inproceedings{li20254dlangsplat,
  title={4D LangSplat: 4D Language Gaussian Splatting via Multimodal Large Language Models},
  author={Wanhua Li and Renping Zhou and Jiawei Zhou and Yingwei Song and Johannes Herter and Minghan Qin and Gao Huang and Hanspeter Pfister},
  booktitle={Proceedings of the IEEE/CVF Conference on Computer Vision and Pattern Recognition},
  year={2025}
}

@inproceedings{gaussian_grouping,
  title={Gaussian Grouping: Segment and Edit Anything in 3D Scenes},
  author={Ye, Mingqiao and Danelljan, Martin and Yu, Fisher and Ke, Lei},
  booktitle={ECCV},
  year={2024}
}

@inproceedings{lerf2023,
  author = {Kerr, Justin and Kim, Chung Min and Goldberg, Ken and Kanazawa, Angjoo and Tancik, Matthew},
  title = {LERF: Language Embedded Radiance Fields},
  booktitle = {International Conference on Computer Vision (ICCV)},
  year = {2023},
}

@inproceedings{charatan2024pixelsplat,
  title={pixelsplat: 3d gaussian splats from image pairs for scalable generalizable 3d reconstruction},
  author={Charatan, David and Li, Sizhe Lester and Tagliasacchi, Andrea and Sitzmann, Vincent},
  booktitle={Proceedings of the IEEE/CVF conference on computer vision and pattern recognition},
  year={2024}
}

@inproceedings{chen2024mvsplat,
  title={Mvsplat: Efficient 3d gaussian splatting from sparse multi-view images},
  author={Chen, Yuedong and Xu, Haofei and Zheng, Chuanxia and Zhuang, Bohan and Pollefeys, Marc and Geiger, Andreas and Cham, Tat-Jen and Cai, Jianfei},
  booktitle={European Conference on Computer Vision},
  year={2024},
}

@inproceedings{szymanowicz2024splatter,
  title={Splatter image: Ultra-fast single-view 3d reconstruction},
  author={Szymanowicz, Stanislaw and Rupprecht, Chrisitian and Vedaldi, Andrea},
  booktitle={Proceedings of the IEEE/CVF conference on computer vision and pattern recognition},
  year={2024}
}

@article{smart2024splatt3r,
  title={Splatt3r: Zero-shot gaussian splatting from uncalibrated image pairs},
  author={Smart, Brandon and Zheng, Chuanxia and Laina, Iro and Prisacariu, Victor Adrian},
  journal={arXiv preprint arXiv:2408.13912},
  year={2024}
}

@inproceedings{clip,
  title={Learning transferable visual models from natural language supervision},
  author={Radford, Alec and Kim, Jong Wook and Hallacy, Chris and Ramesh, Aditya and Goh, Gabriel and Agarwal, Sandhini and Sastry, Girish and Askell, Amanda and Mishkin, Pamela and Clark, Jack and others},
  booktitle={International conference on machine learning},
  year={2021},
}

@article{sun2025uni3r,
  title={Uni3R: Unified 3D Reconstruction and Semantic Understanding via Generalizable Gaussian Splatting from Unposed Multi-View Images},
  author={Xiangyu Sun and Haoyi Jiang and Liu Liu and Seungtae Nam and Gyeongjin Kang and Xinjie Wang and Wei Sui and Zhizhong Su and Wenyu Liu and Xinggang Wang and Eunbyung Park},
  year={2025},
  journal={arXiv preprint arXiv:2508.03643},
}

@inproceedings{peng2023openscene,
  author    = {Peng, Songyou and Genova, Kyle and Jiang, Chiyu {\textquotedblleft}Max{\textquotedblright} and Tagliasacchi, Andrea and Pollefeys, Marc and Funkhouser, Thomas},
  title     = {OpenScene: 3D Scene Understanding With Open Vocabularies},
  booktitle = {Proceedings of the IEEE/CVF Conference on Computer Vision and Pattern Recognition (CVPR)},
  year      = {2023},
}

@article{tian2025fleg,
  title={FLEG: Feed-Forward Language Embedded Gaussian Splatting from Any Views},
  author={Qijian Tian and Xin Tan and Jiayu Ying and Xuhong Wang and Yuan Xie and Lizhuang Ma},
  year={2025},
  journal={arXiv preprint arXiv:2512.17541},
}

@article{tian2025uniforward,
  title={UniForward: Unified 3D Scene and Semantic Field Reconstruction via Feed-Forward Gaussian Splatting from Only Sparse-View Images},
  author={Qijian Tian and Xin Tan and Jingyu Gong and Yuan Xie and Lizhuang Ma},
  year={2025},
  journal={arXiv preprint arXiv:2506.09378},
}

@article{fan2024lsm,
  title={Large spatial model: End-to-end unposed images to semantic 3d},
  author={Fan, Zhiwen and Zhang, Jian and Cong, Wenyan and Wang, Peihao and Li, Renjie and Wen, Kairun and Zhou, Shijie and Kadambi, Achuta and Wang, Zhangyang and Xu, Danfei and others},
  journal={Advances in neural information processing systems},
  year={2024}
}

@article{sheng2025spatialsplat,
  title={SpatialSplat: Efficient Semantic 3D from Sparse Unposed Images},
  author={Sheng, Yu and Deng, Jiajun and Zhang, Xinran and Zhang, Yu and Hua, Bei and Zhang, Yanyong and Ji, Jianmin},
  journal={arXiv preprint arXiv:2505.23044},
  year={2025}
}

@inproceedings{li2024hierarchical,
  title={Hierarchical temporal context learning for camera-based semantic scene completion},
  author={Li, Bohan and Deng, Jiajun and Zhang, Wenyao and Liang, Zhujin and Du, Dalong and Jin, Xin and Zeng, Wenjun},
  booktitle={European Conference on Computer Vision},
  year={2024},
}

@InProceedings{huang2024selfocc,
  author    = {Huang, Yuanhui and Zheng, Wenzhao and Zhang, Borui and Zhou, Jie and Lu, Jiwen},
  title     = {SelfOcc: Self-Supervised Vision-Based 3D Occupancy Prediction},
  booktitle = {Proceedings of the IEEE/CVF Conference on Computer Vision and Pattern Recognition (CVPR)},
  year      = {2024},
}

@InProceedings{wu2025embodiedocc,
  author    = {Wu, Yuqi and Zheng, Wenzhao and Zuo, Sicheng and Huang, Yuanhui and Zhou, Jie and Lu, Jiwen},
  title     = {EmbodiedOcc: Embodied 3D Occupancy Prediction for Vision-based Online Scene Understanding},
  booktitle = {Proceedings of the IEEE/CVF International Conference on Computer Vision (ICCV)},
  year      = {2025},
}

@article{wang2025volsplat,
  title={VolSplat: Rethinking Feed-Forward 3D Gaussian Splatting with Voxel-Aligned Prediction},
  author={Wang, Weijie and Chen, Yeqing and Zhang, Zeyu and Liu, Hengyu and Wang, Haoxiao and Feng, Zhiyuan and Qin, Wenkang and Zhu, Zheng and Chen, Donny Y. and Zhuang, Bohan},
  journal={arXiv preprint arXiv:2509.19297},
  year={2025}
}

@inproceedings{keetha2026mapanything,
  title={MapAnything: Universal Feed-Forward Metric 3D Reconstruction},
  author={Nikhil Keetha and Norman M\"{u}ller and Johannes Sch\"{o}nberger and Lorenzo Porzi and Yuchen Zhang and Tobias Fischer and Arno Knapitsch and Duncan Zauss and Ethan Weber and Nelson Antunes and Jonathon Luiten and Manuel Lopez-Antequera and Samuel Rota Bul\`{o} and Christian Richardt and Deva Ramanan and Sebastian Scherer and Peter Kontschieder},
  booktitle={International Conference on 3D Vision (3DV)},
  year={2026},
}

@inproceedings{konushin2025tun3d,
  title={TUN3D: Towards Real-World Scene Understanding from Unposed Images},
  author={Anton Konushin and Nikita Drozdov and Bulat Gabdullin and Alexey Zakharov and Anna Vorontsova and Danila Rukhovich and Maksim Kolodiazhnyi},
  year={2025},
  booktitle = {IEEE International Conference on Robotics and Automation (ICRA)},
}

@inproceedings{epipolartransformers,
  title={Epipolar Transformers},
  author={He, Yihui and Yan, Rui and Fragkiadaki, Katerina and Yu, Shoou-I},
  booktitle={Proceedings of the IEEE/CVF Conference on Computer Vision and Pattern Recognition},
  year={2020}
}

@article{yao2018mvsnet,
  title={MVSNet: Depth Inference for Unstructured Multi-view Stereo},
  author={Yao, Yao and Luo, Zixin and Li, Shiwei and Fang, Tian and Quan, Long},
  journal={European Conference on Computer Vision (ECCV)},
  year={2018}
}

@article{chen2024mvsplat360,
  title     = {MVSplat360: Feed-Forward 360 Scene Synthesis from Sparse Views},
  author    = {Chen, Yuedong and Zheng, Chuanxia and Xu, Haofei and Zhuang, Bohan and Vedaldi, Andrea and Cham, Tat-Jen and Cai, Jianfei},
  journal = {Advances in Neural Information Processing Systems (NeurIPS)},
  year      = {2024},
}

@inproceedings{zhang2025flare,
  title={Flare: Feed-forward geometry, appearance and camera estimation from uncalibrated sparse views},
  author={Zhang, Shangzhan and Wang, Jianyuan and Xu, Yinghao and Xue, Nan and Rupprecht, Christian and Zhou, Xiaowei and Shen, Yujun and Wetzstein, Gordon},
  booktitle={Proceedings of the Computer Vision and Pattern Recognition Conference},
  year={2025}
}

@inproceedings{hong2025pfplat,
  title={PF3plat: Pose-Free Feed-Forward 3D Gaussian Splatting for Novel View Synthesis},
  author={Sunghwan Hong and Jaewoo Jung and Heeseong Shin and Jisang Han and Jiaolong Yang and Chong Luo and Seungryong Kim},
  booktitle={Forty-second International Conference on Machine Learning},
  year={2025}
}

@inproceedings{ye2026yonosplat,
  title     = {YoNoSplat: You Only Need One Model for Feedforward 3D Gaussian Splatting},
  author    = {Ye, Botao and Chen, Boqi and Xu, Haofei and Barath, Daniel and Pollefeys, Marc},
  booktitle = {International Conference on Learning Representations (ICLR)},
  year      = {2026}
}

@inproceedings{ye2025noposplat,
  title={No Pose, No Problem: Surprisingly Simple 3D Gaussian Splats from Sparse Unposed Images},
  author={Botao Ye and Sifei Liu and Haofei Xu and Xueting Li and Marc Pollefeys and Ming-Hsuan Yang and Songyou Peng},
  booktitle={International Conference on Learning Representations (ICLR)},
  year={2025},
}

@inproceedings{tian2025drivingforward,
  title={DrivingForward: Feed-forward 3D Gaussian Splatting for Driving Scene Reconstruction from Flexible Surround-view Input},
  author={Qijian Tian and Xin Tan and Yuan Xie and Lizhuang Ma},
  booktitle={Proceedings of the AAAI Conference on Artificial Intelligence},
  year={2025}
}

@article{li2025sliceocc,
  title={SliceOcc: Indoor 3D Semantic Occupancy Prediction with Vertical Slice Representation},
  author={Jianing Li and Ming Lu and Hao Wang and Chenyang Gu and Wenzhao Zheng and Li Du and Shanghang Zhang},
  journal={2025 IEEE International Conference on Robotics and Automation (ICRA)},
  year={2025},
}

@article{wei2023surroundocc,
  title={SurroundOcc: Multi-Camera 3D Occupancy Prediction for Autonomous Driving},
  author={Yi Wei and Linqing Zhao and Wenzhao Zheng and Zhengbiao Zhu and Jie Zhou and Jiwen Lu},
  journal={2023 IEEE/CVF International Conference on Computer Vision (ICCV)},
  year={2023},
}

@inproceedings{tian2023occ3d,
  title={Occ3D: A Large-Scale 3D Occupancy Prediction Benchmark for Autonomous Driving},
  author={Xiaoyu Tian and Tao Jiang and Longfei Yun and Yucheng Mao and Huitong Yang and Yue Wang and Yilun Wang and Hang Zhao},
  booktitle={Thirty-seventh Conference on Neural Information Processing Systems Datasets and Benchmarks Track},
  year={2023},
}

@article{wang2023openoccupancy,
  title={OpenOccupancy: A Large Scale Benchmark for Surrounding Semantic Occupancy Perception},
  author={Xiaofeng Wang and Zhengbiao Zhu and Wenbo Xu and Yunpeng Zhang and Yi Wei and Xu Chi and Yun Ye and Dalong Du and Jiwen Lu and Xingang Wang},
  journal={2023 IEEE/CVF International Conference on Computer Vision (ICCV)},
  year={2023},
}

@article{behley2019semantickitti,
  title={SemanticKITTI: A Dataset for Semantic Scene Understanding of LiDAR Sequences},
  author={Jens Behley and Martin Garbade and Andres Milioto and Jan Quenzel and Sven Behnke and C. Stachniss and Juergen Gall},
  journal={2019 IEEE/CVF International Conference on Computer Vision (ICCV)},
  year={2019},
}

@article{li2023voxformer,
  title={VoxFormer: Sparse Voxel Transformer for Camera-Based 3D Semantic Scene Completion},
  author={Yiming Li and Zhiding Yu and Christopher Bongsoo Choy and Chaowei Xiao and Jos{\'e} Manuel {\'A}lvarez and Sanja Fidler and Chen Feng and Anima Anandkumar},
  journal={2023 IEEE/CVF Conference on Computer Vision and Pattern Recognition (CVPR)},
  year={2023},
}

@article{huang2023tpvformer,
  title={Tri-Perspective View for Vision-Based 3D Semantic Occupancy Prediction},
  author={Yuan-Ko Huang and Wenzhao Zheng and Yunpeng Zhang and Jie Zhou and Jiwen Lu},
  journal={2023 IEEE/CVF Conference on Computer Vision and Pattern Recognition (CVPR)},
  year={2023},
}

@article{zhang2023occformer,
  title={OccFormer: Dual-path Transformer for Vision-based 3D Semantic Occupancy Prediction},
  author={Yunpeng Zhang and Zhengbiao Zhu and Dalong Du},
  journal={IEEE/CVF International Conference on Computer Vision (ICCV)},
  year={2023},
}

@article{huang2024gaussianformer2,
  title={GaussianFormer-2: Probabilistic Gaussian Superposition for Efficient 3D Occupancy Prediction},
  author={Yuanhui Huang and Amonnut Thammatadatrakoon and Wenzhao Zheng and Yunpeng Zhang and Dalong Du and Jiwen Lu},
  journal={IEEE/CVF Conference on Computer Vision and Pattern Recognition (CVPR)},
  year={2025},
}

@article{huang2024gaussianformer,
  author = {Huang, Yuanhui and Zheng, Wenzhao and Zhang, Yunpeng and Zhou, Jie and Lu, Jiwen},
  title = {GaussianFormer: Scene as Gaussians for Vision-Based 3D Semantic Occupancy Prediction},
  year = {2024},
  journal={European Conference on Computer Vision (ECCV)},
}

@article{chen2025casualocc,
  title={Semantic Causality-Aware Vision-Based 3D Occupancy Prediction},
  author={Dubing Chen and Huan Zheng and Yucheng Zhou and Xianfei Li and Wenlong Liao and Tao He and Pai Peng and Jianbing Shen},
  journal={IEEE/CVF International Conference on Computer Vision (ICCV)},
  year={2025}
}

@article{gan2024gaussianocc,
  title={GaussianOcc: Fully Self-supervised and Efficient 3D Occupancy Estimation with Gaussian Splatting},
  author={Wanshui Gan and Fang Liu and Hongbin Xu and Ningkai Mo and Naoto Yokoya},
  year={2025},
  journal={IEEE/CVF International Conference on Computer Vision (ICCV)},

}

@article{gan2024simpleocc,
  title={A Comprehensive Framework for 3D Occupancy Estimation in Autonomous Driving},
  author={Gan, Wanshui and Mo, Ningkai and Xu, Hongbin and Yokoya, Naoto},
  journal={IEEE Transactions on Intelligent Vehicles},
  year={2024},
}

@article{zhang2025occnerf,
  title={Occnerf: Advancing 3d occupancy prediction in lidar-free environments},
  author={Zhang, Chubin and Yan, Juncheng and Wei, Yi and Li, Jiaxin and Liu, Li and Tang, Yansong and Duan, Yueqi and Lu, Jiwen},
  journal={IEEE Transactions on Image Processing},
  year={2025},
}

@inproceedings{wang2025embodiedocc_pp,
  title={EmbodiedOcc++: Boosting Embodied 3D Occupancy Prediction with Plane Regularization and Uncertainty Sampler},
  author={Hao Wang and Xiaobao Wei and Xiaoan Zhang and Jianing Li and Chengyu Bai and Ying Li and Ming Lu and Wenzhao Zheng and Shanghang Zhang},
  booktitle={Proceedings of the 33rd ACM International Conference on Multimedia},
  year={2025},
}

@article{xuepinet2026,
  author={Xue, Yujie and Pi, Huilong and Tang, Zhuo and Li, Kenli and Li, Ruihui},
  journal={IEEE Transactions on Multimedia}, 
  title={PI-Net: Point-to-Image Knowledge Distillation for Camera-based 3D Semantic Scene Completion}, 
  year={2026},
  doi={10.1109/TMM.2026.3668494}
  }

@article{jia2022conv,
  author={Jia, Wei and Li, Li and Akhtar, Anique and Li, Zhu and Liu, Shan},
  journal={IEEE Transactions on Multimedia}, 
  title={Convolutional Neural Network-Based Occupancy Map Accuracy Improvement for Video-Based Point Cloud Compression}, 
  year={2022},
  doi={10.1109/TMM.2021.3079698}}

@article{wang2026adapt,
  author={Wang, Changwei and Xu, Wenhao and Xu, Rongtao and Zhang, Zherui and Xu, Shibiao and Zhang, Jiguang and Teng, Xiaoqiang and Meng, Weiliang and Zhang, Xiaopeng},
  journal={IEEE Transactions on Multimedia}, 
  title={Adaptive in Adapter: Boosting Open-Vocabulary Semantic Segmentation With Adaptive Dropout Adapter}, 
  year={2026},
  doi={10.1109/TMM.2026.3654453}}

@article{zhang2026detagent,
  author={Zhang, Ruisong and Wu, Xin-Jian and Wang, Chuang and Liu, Cheng-Lin},
  journal={IEEE Transactions on Multimedia}, 
  title={Det-Agent: Open-Vocabulary Object Localization and Detection with Reinforcement Learning Agent}, 
  year={2026},
  doi={10.1109/TMM.2026.3668516}}

@article{zhang2025unleash,
  author={Zhang, Wenyao and Wu, Letian and Zhang, Zequn and Yu, Tao and Ma, Chao and Jin, Xin and Yang, Xiaokang and Zeng, Wenjun},
  journal={IEEE Transactions on Multimedia}, 
  title={Unleash the Power of Vision-Language Models by Visual Attention Prompt and Multimodal Interaction}, 
  year={2025},
  doi={10.1109/TMM.2024.3521785}}

@article{li2023from,
  author={Li, Jie and Song, Qi and Yan, Xiaohu and Chen, Yongquan and Huang, Rui},
  journal={IEEE Transactions on Multimedia}, 
  title={From Front to Rear: 3D Semantic Scene Completion Through Planar Convolution and Attention-Based Network}, 
  year={2023},
  doi={10.1109/TMM.2023.3234441}}

@article{zhang20253dgeo,
  author={Zhang, Yi and Wang, Yi and Cui, Yawen and Chau, Lap-Pui},
  journal={IEEE Transactions on Multimedia}, 
  title={3DGeoDet: General-Purpose Geometry-Aware Image-Based 3D Object Detection}, 
  year={2025},
  doi={10.1109/TMM.2025.3581780}}

@article{xiong20253urllm,
  author={Xiong, Haomiao and Zhuge, Yunzhi and Zhu, Jiawen and Zhang, Lu and Lu, Huchuan},
  journal={IEEE Transactions on Multimedia}, 
  title={3UR-LLM: An End-to-End Multimodal Large Language Model for 3D Scene Understanding}, 
  year={2025},
  doi={10.1109/TMM.2025.3557620}}

@article{fang2026casovd,
  author={Fang, Zhenyu and Wu, Yulong and Ren, Jinchang and Zheng, Jiangbin and Yan, Yijun and Zhang, Lixiang},
  journal={IEEE Transactions on Multimedia}, 
  title={Cas-OVD: Cascaded Open-Vocabulary Detection of Small Objects Using Multi-Refined Region Proposal Network in Autonomous Driving}, 
  year={2026},
  doi={10.1109/TMM.2025.3632649}}

@article{li2025omninwm,
  title={OmniNWM: Omniscient Driving Navigation World Models},
  author={Li, Bohan and Ma, Zhuang and Du, Dalong and Peng, Baorui and Liang, Zhujin and Liu, Zhenqiang and Ma, Chao and Jin, Yueming and Zhao, Hao and Zeng, Wenjun and others},
  journal={arXiv preprint arXiv:2510.18313},
  year={2025}
}

@inproceedings{li2025uniscene,
  title={Uniscene: Unified occupancy-centric driving scene generation},
  author={Li, Bohan and Guo, Jiazhe and Liu, Hongsi and Zou, Yingshuang and Ding, Yikang and Chen, Xiwu and Zhu, Hu and Tan, Feiyang and Zhang, Chi and Wang, Tiancai and others},
  booktitle={Proceedings of the computer vision and pattern recognition conference},
  pages={11971--11981},
  year={2025}
}

@article{li2026scaling,
  title={Scaling up occupancy-centric driving scene generation: Dataset and method},
  author={Li, Bohan and Jin, Xin and Zhu, Hu and Liu, Hongsi and Li, Ruikai and Guo, Jiazhe and Cai, Kaiwen and Ma, Chao and Jin, Yueming and Zhao, Hao and others},
  journal={IEEE Transactions on Pattern Analysis and Machine Intelligence},
  year={2026},
  publisher={IEEE}
}

@article{li2025occscene,
  title={OccScene: Semantic occupancy-based cross-task mutual learning for 3D scene generation},
  author={Li, Bohan and Jin, Xin and Wang, Jianan and Shi, Yukai and Sun, Yasheng and Wang, Xiaofeng and Ma, Zhuang and Xie, Baao and Ma, Chao and Yang, Xiaokang and others},
  journal={IEEE Transactions on Pattern Analysis and Machine Intelligence},
  year={2025},
  publisher={IEEE}
}

@article{li2026hierarchical,
  title={Hierarchical Context Alignment with Disentangled Geometric and Temporal Modeling for Semantic Occupancy Prediction},
  author={Li, Bohan and Deng, Jiajun and Sun, Yasheng and Wang, Xiaofeng and Jin, Xin and Zeng, Wenjun},
  journal={IEEE Transactions on Pattern Analysis and Machine Intelligence},
  year={2026},
  publisher={IEEE}
}

@article{li2026articulated,
  title={From Articulated Kinematics to Routed Visual Control for Action-Conditioned Surgical Video Generation},
  author={Li, Bohan and Yang, Shuojue and Peng, Baorui and Guo, Xianda and Zhang, Erli and Tao, Youqi and Duan, Junfeng and Xu, Daguang and Dou, Qi and Jin, Xin and others},
  journal={arXiv preprint arXiv:2605.08712},
  year={2026}
}

@article{li2023bridging,
  title={Bridging stereo geometry and BEV representation with reliable mutual interaction for semantic scene completion},
  author={Li, Bohan and Sun, Yasheng and Liang, Zhujin and Du, Dalong and Zhang, Zhuanghui and Wang, Xiaofeng and Wang, Yunnan and Jin, Xin and Zeng, Wenjun},
  journal={arXiv preprint arXiv:2303.13959},
  year={2023}
}

@inproceedings{li2024one,
  title={One at a time: Progressive multi-step volumetric probability learning for reliable 3d scene perception},
  author={Li, Bohan and Sun, Yasheng and Dong, Jingxin and Zhu, Zheng and Liu, Jinming and Jin, Xin and Zeng, Wenjun},
  booktitle={Proceedings of the AAAI Conference on Artificial Intelligence},
  volume={38},
  number={4},
  pages={3028--3036},
  year={2024}
}

\vfill

\end{document}